\documentclass[html,twocolumn]{article}
\usepackage{amsmath}
\usepackage{graphicx}
\usepackage{algorithmic}
\usepackage{algorithm}
\usepackage{hyperref}
\usepackage{verbatim}
\usepackage{enumerate}
\usepackage{amsfonts}
\usepackage{amsthm, amssymb}
\usepackage{epstopdf}
\usepackage{subfig}
\usepackage{color}
\usepackage{tabu}
\usepackage{multirow}
\usepackage[table]{xcolor}
\graphicspath{{./images/}}

\title{Depth Perception in Autostereograms: 1/f-Noise is Best}

\usepackage{authblk}
\author{Yael Yankelevsky, Ishai Shvartz, Tamar Avraham and Alfred M. Bruckstein}
\affil{Technion - I.I.T, Haifa 32000, Israel}
\date{}

\begin{document}

\maketitle

\begin{abstract}
An autostereogram is a single image that encodes depth information that pops out when looking at it. The trick is achieved by replicating a vertical strip that sets a basic two-dimensional pattern with disparity shifts that encode a three-dimensional scene. 
It is of interest to explore the dependency between the ease of perceiving depth in autostereograms and the choice of the basic pattern used for generating them.
In this work we confirm a theory proposed by \cite{Bruckstein1996} to explain the process of autostereographic depth perception, providing a measure for the ease of ``locking into" the depth profile, based on the spectral properties of the basic pattern used. 
We report the results of three sets of psychophysical experiments using autostereograms generated from two-dimensional random noise patterns having power spectra of the form $1/f^\beta$.
The experiments were designed to test the ability of human subjects to identify smooth, low resolution surfaces, as well as detail, in the form of higher resolution objects in the depth profile, and to determine limits in identifying small objects as a function of their size. In accordance with the theory, we discover a significant advantage of the $1/f$ noise pattern (pink noise) for fast depth lock-in and fine detail detection, showing that such patterns are optimal choices for autostereogram design.
Validating the theoretical model predictions strengthens its underlying assumptions, and contributes to a better understanding of the visual system's binocular disparity mechanisms.
\end{abstract}

\section{Introduction} \label{section:Introduction}
While the world around us is three-dimensional, the visual data is inherently two-dimensional. Nevertheless, the third dimension can often be inferred from one or more images, utilizing cues such as occlusion, size, texture, lighting and shading, prior shape information etc. \cite{Palmer1999}.
Stereo vision provides binocular cues such as the vergence angle, formed between the axes from the eyes to the convergence point on which both eyes fixate, and, most importantly, binocular disparity, which is the horizontal displacement between matching features in the images acquired by pairs of eyes (or cameras).

The seminal work of Julesz \cite{Julesz1964} established that depth information can be retrieved by determining correspondences of matching local features.
The correspondence problem is widely studied and over the years many 
models for stereopsis have been proposed (e.g.~\cite{Tyler1977,Marr1979,Sperling1981,Li1994,Arndt1995,Reimann1995,
Qian1997,Harris1997,Banks2004,Read2007,Burge2014}).
Julesz‎ \cite{Julesz1964} further showed that depth can be perceived from disparity alone, without any other visual cues, by creating \emph{Random-Dot-Stereograms}, which are pairs of similar images consisting of randomly placed dots in the plane, one of them having part of the dots slightly displaced to encode depth.

The phenomenon of seeing illusory depth in repeating patterns was revealed even earlier by Brewster \cite{Brewster1844} in what became known as the \emph{wallpaper effect}. Ittelson \cite{Ittelson1960} also reported that effect, observing that wall surfaces with repeating patterns appear to move forward after long stares, and letters on a typewriter keyboard sometimes perceptually merge into one. 

Tyler et al. \cite{Tyler1977,Tyler1990} further studied this idea and discovered that depth can also be encoded in a single random dot image by cleverly producing ``stereo pairs" that are identical, hence effectively fusing them into one image, called a \emph{Single-Image Random-Dot Stereogram} (SIRDS) or more generally an \emph{Autostereogram}.

An autostereogram is a periodic pattern with horizontal deformations modulated by a spatial depth function, such that pixel values are a function of the constant distance between the eyes and the encoded scene's local depth. Corresponding pixels, i.e. pixels which originate at the same point in the depth scene, are given the same color or gray-level value, and being seen identical by the two eyes, they will potentially be matched in the binocular viewing process.

Autostereograms became widespread as a very popular art form due to a series of books titled ``Magic Eye" \cite{Magic1993} and were proposed for applications like 3D photography and computer graphics. 
Since they can be viewed without special equipment and are easily manipulated to create various visual effects and illusions, autostereograms can serve as an important tool in the study of depth perception and in binocular vision research in general (see e.g. \cite{Wilmer2008,Gomez2012}).

As part of the research efforts on autostereograms, Tyler and Clarke \cite{Tyler1990} also invented more complex autostereograms capable of encoding multiple depths. This concept was further examined by \cite{Terrell1994}.

Thimbleby et al. \cite{Thimbleby1994} proposed a computational algorithm for autostereograms generation based on geometrical constraints. Over the years, other algorithms for autostereogram generation were suggested such as those by Minh et al. \cite{Minh2001,Minh2002} and by Geselowitz \cite{Geselowitz2003}.
Some research also dealt with depth-map reconstruction from autostereograms \cite{Kimmel2002,Lau2002}. 

Only few studies investigated the conditions for which autostereogram viewing is easier. Ditzinger et al. \cite{Ditzinger2000} found that random noise added to the repeated patterns can improve depth perception and reduce hysteresis effects.
The work of Bruckstein et al. \cite{Bruckstein1996} is the only one to deal with the effect of the choice of basic pattern and to develop a theoretical framework for analyzing the ease of depth perception in autostereograms with respect to the underlying noise pattern used for the autostereogram generation.
Their model predicts that autostereograms created from pink noise patterns (having a power spectrum of $1/f^\beta \; ; \;\beta=1$) should be more easily perceived than those created using other noise patterns, as detailed in the Appendix.
Whereas depth ``lock-in" is anticipated to break at fine scales for $\beta>1$ and at coarse scales for $\beta<1$, pink noise leads to scale invariant match functions and is therefore optimal for locking in and maintaining the depth perception effect across scales.

This theoretical result nicely resonates with studies on natural images, that found that such images tend to have power spectra of $1/f$ \cite{Burton1987,Torralba2003}, suggesting that our eyes may be adapted by evolution to optimally perceive such patterns.

Though mathematically well established, the model suggested by \cite{Bruckstein1996} has never been experimentally tested.
The current work aims to further understand the autostereogram depth perception mechanism by testing and verifying this model using psychophysical experiments.

\section{Methods} \label{section:Methods}

In order to test and confirm the prediction of the model suggested by \cite{Bruckstein1996}, we conducted three sets of psychophysical experiments using autostereograms.
In the first two experiments we evaluated the effect of different noise patterns on perception of low resolution surfaces and higher resolution objects (letter profiles) in the depth dimension. In the third experiment we explored the effect of different noise patterns on the limits of identifying small objects in the depth dimension as a function of their size.

\subsection{Subjects}
Fifteen participants between the ages of 20 and 34 took part in the experiments.
All the participants had normal or corrected-to-normal vision, and all were first tested for their ability to perceive depth hidden in autostereograms.
The participants were unaware of the purpose of the study.
In the third experiment, only seven of the original fifteen participants took part.

\subsection{Stimuli}
The stimuli in all our experiments are autostereograms that were generated from pre-designed depth maps and a desired noise spectrum according to the algorithm described below:

First, a basic noise patch of $128\times 128$ pixels having a noise spectrum of $1/f^\beta$ was stochastically generated for each autostereogram. 
The patch was then up-sampled by a factor of 2 to create a $256\times 256$ block with basic ``pixels" of size $2\times 2$.
Up-sampling was performed to enable easier viewing of the autostereograms on large monitors.
The up-sampled patch was replicated vertically to create a strip of $256\times 1024$ pixels, and the strip was horizontally replicated 6 times based on the viewing geometry and the given depth map, similarly to the process described in \cite{Thimbleby1994}, resulting in the final autostereogram sized $1536\times 1024$.

Next, we shall describe the depth maps designed for each experiment.

\subsubsection{Experiment 1 - Surface Recognition}
\begin{figure}[!htbp] 
\centering
\subfloat[]{
\centering \includegraphics[scale=0.13,clip,trim=0cm 0cm 0cm 0cm]{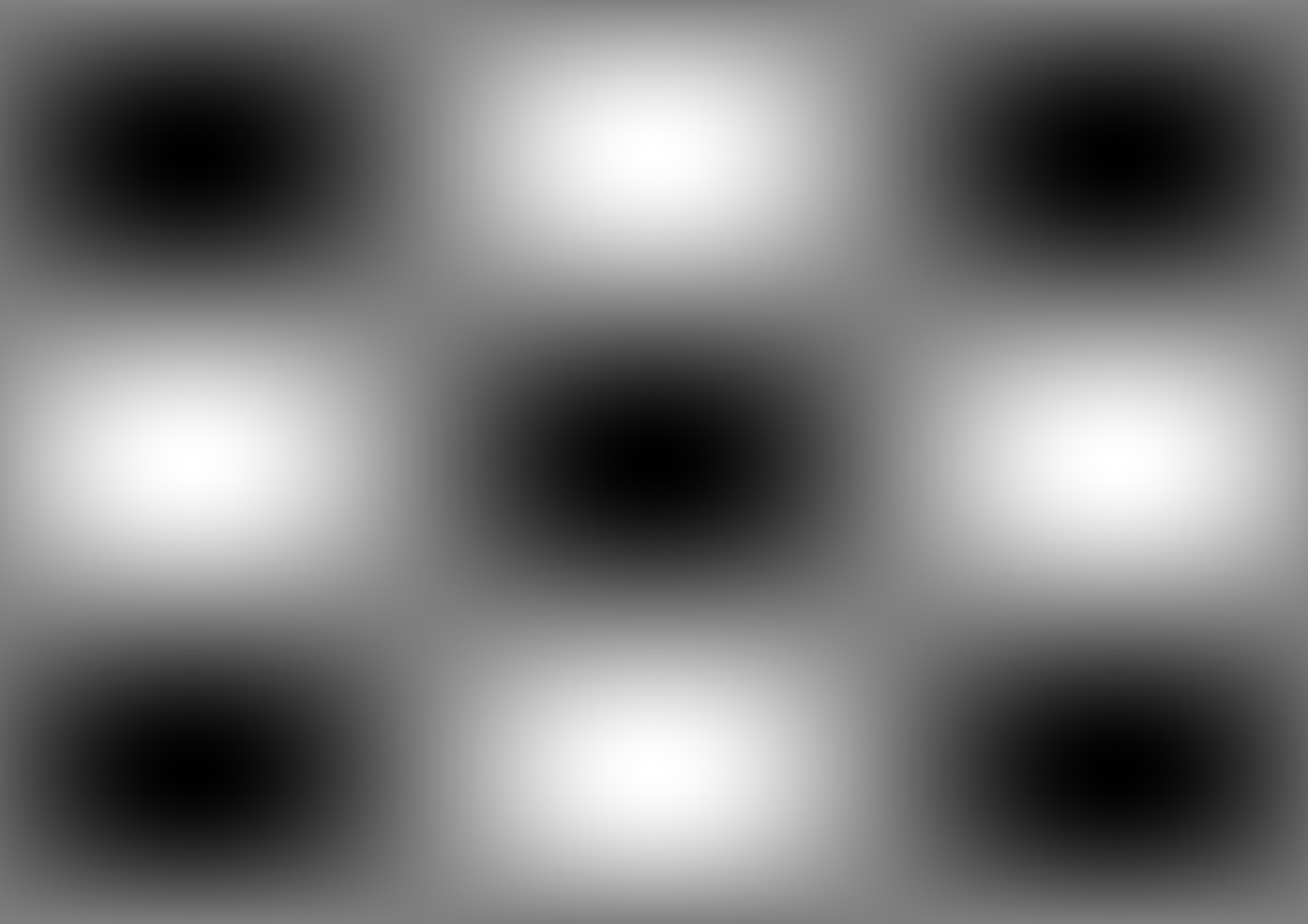}
\label{Fig:eggcrate_depthmap}
}
\subfloat[]{
\centering \includegraphics[scale=0.13,clip,trim=0cm 0cm 0cm 0cm]{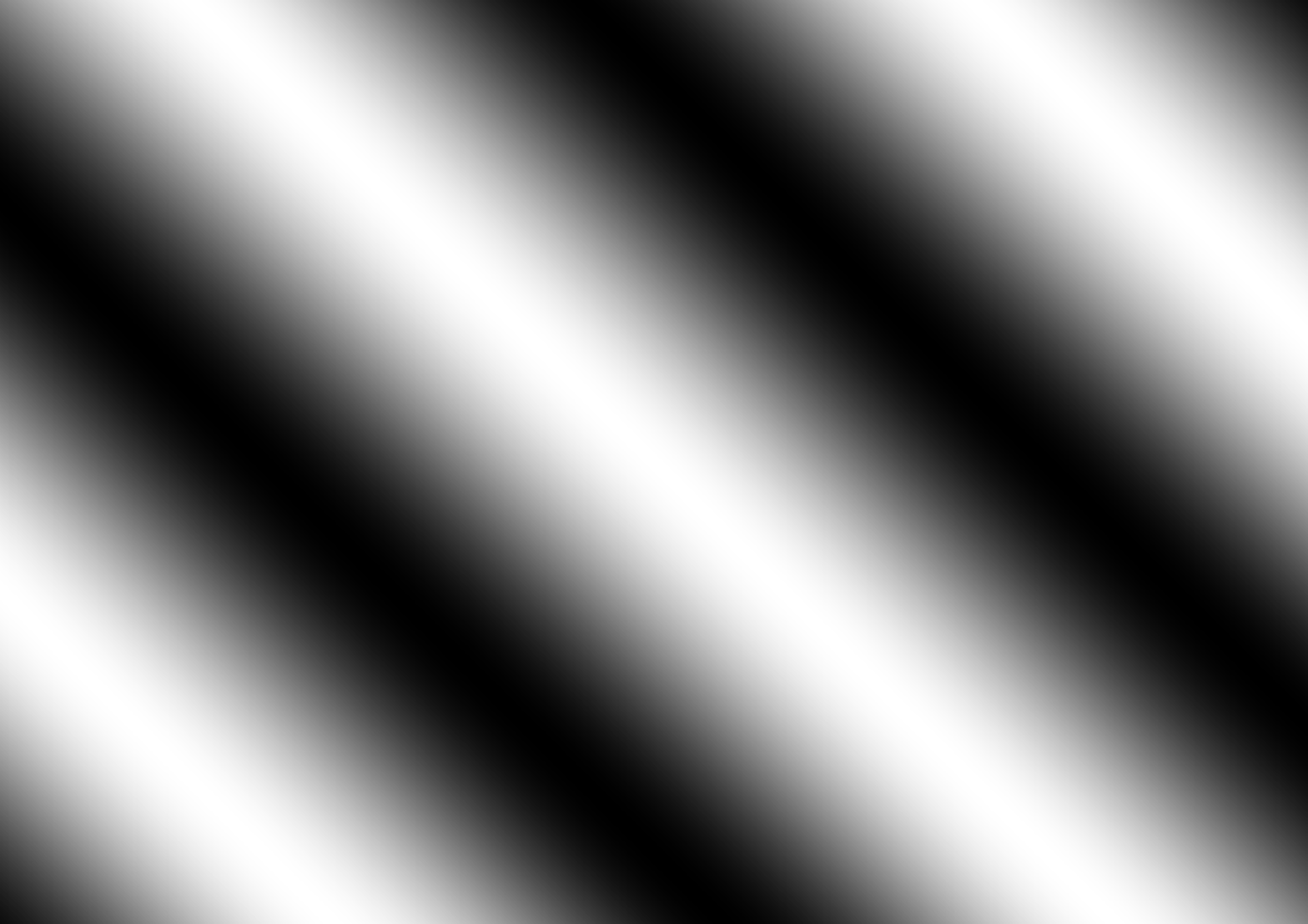}
\label{Fig:diag_depthmap}
}\\
\centering
\subfloat[]{
\centering \includegraphics[scale=0.13,clip,trim=0cm 0cm 0cm 0cm]{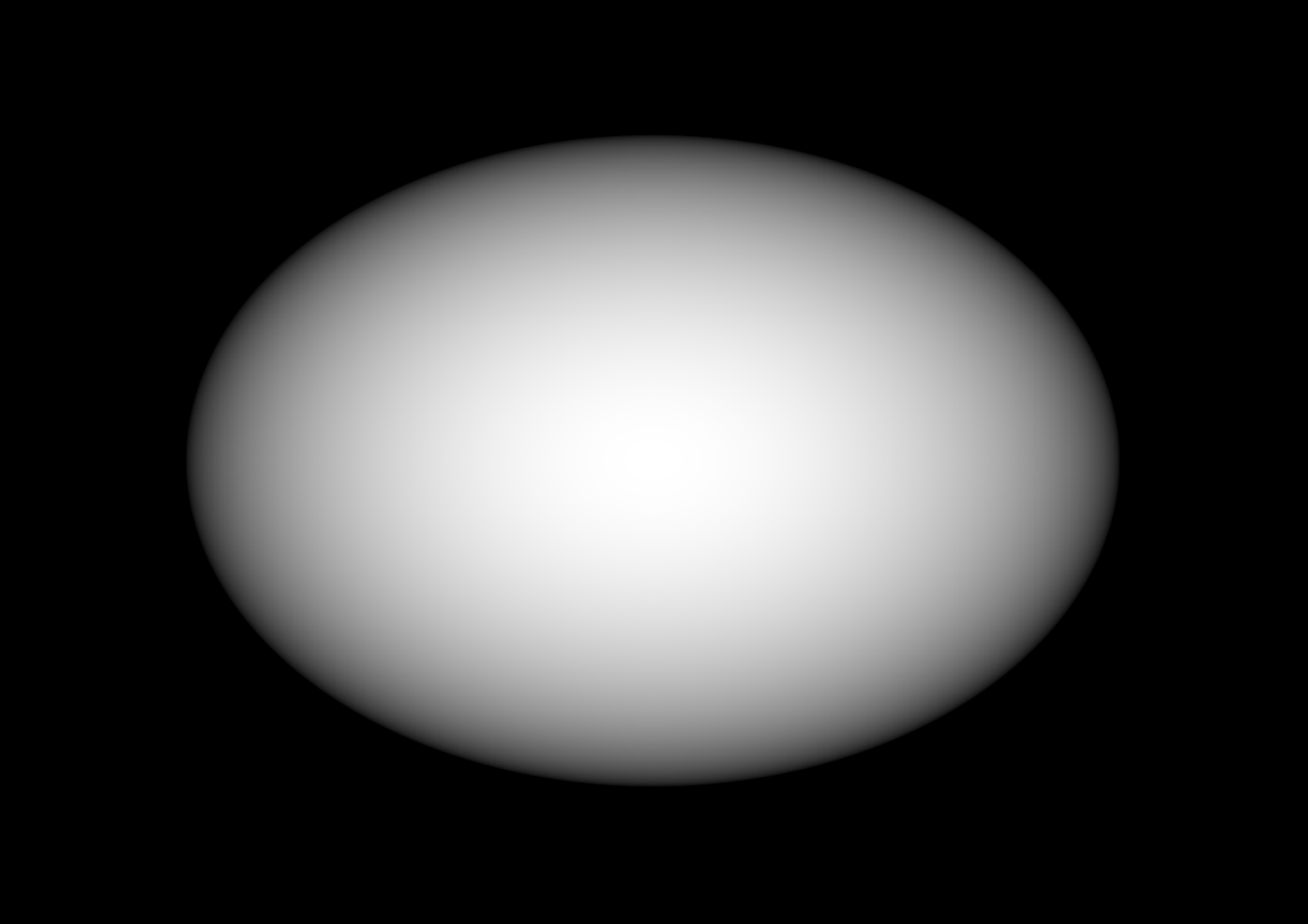}
\label{Fig:sphere_depthmap}
}
\subfloat[]{
\centering \includegraphics[scale=0.13,clip,trim=0cm 0cm 0cm 0cm]{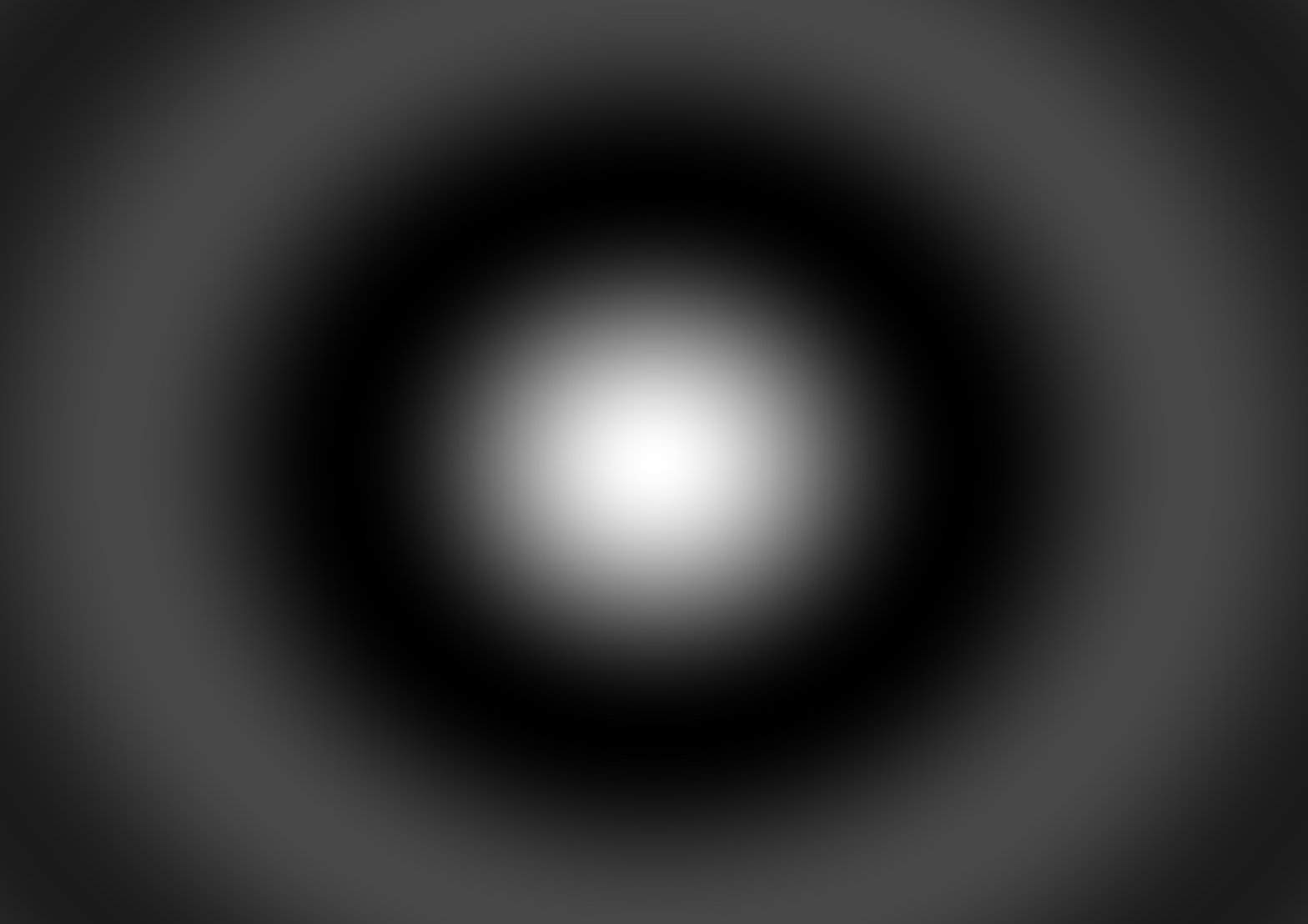}
\label{Fig:mexican_depthmap}
}
\caption[]{Depth maps used for experiment 1: \subref{Fig:eggcrate_depthmap} Egg crate, \subref{Fig:diag_depthmap} Diagonal sinus wave, \subref{Fig:sphere_depthmap} Ellipsoid, \subref{Fig:mexican_depthmap} Mexican hat}
\label{Fig:DepthMaps}
\end{figure}

Low resolution in the depth dimension was represented by depth maps of four different smooth profiles.
These depth profiles were designed using continuous functions so as to avoid occlusions, miss-correspondences and echoes in the creation of the autostereogram \cite{Thimbleby1994} and were considerably different from one another in order to avoid confusion in their identification (see Figure~\ref{Fig:DepthMaps}).

\begin{figure*}[!htbp] 
\centering
\subfloat[]{
\centering \includegraphics[scale=0.23,clip,trim=0cm 0cm 0cm 0cm]{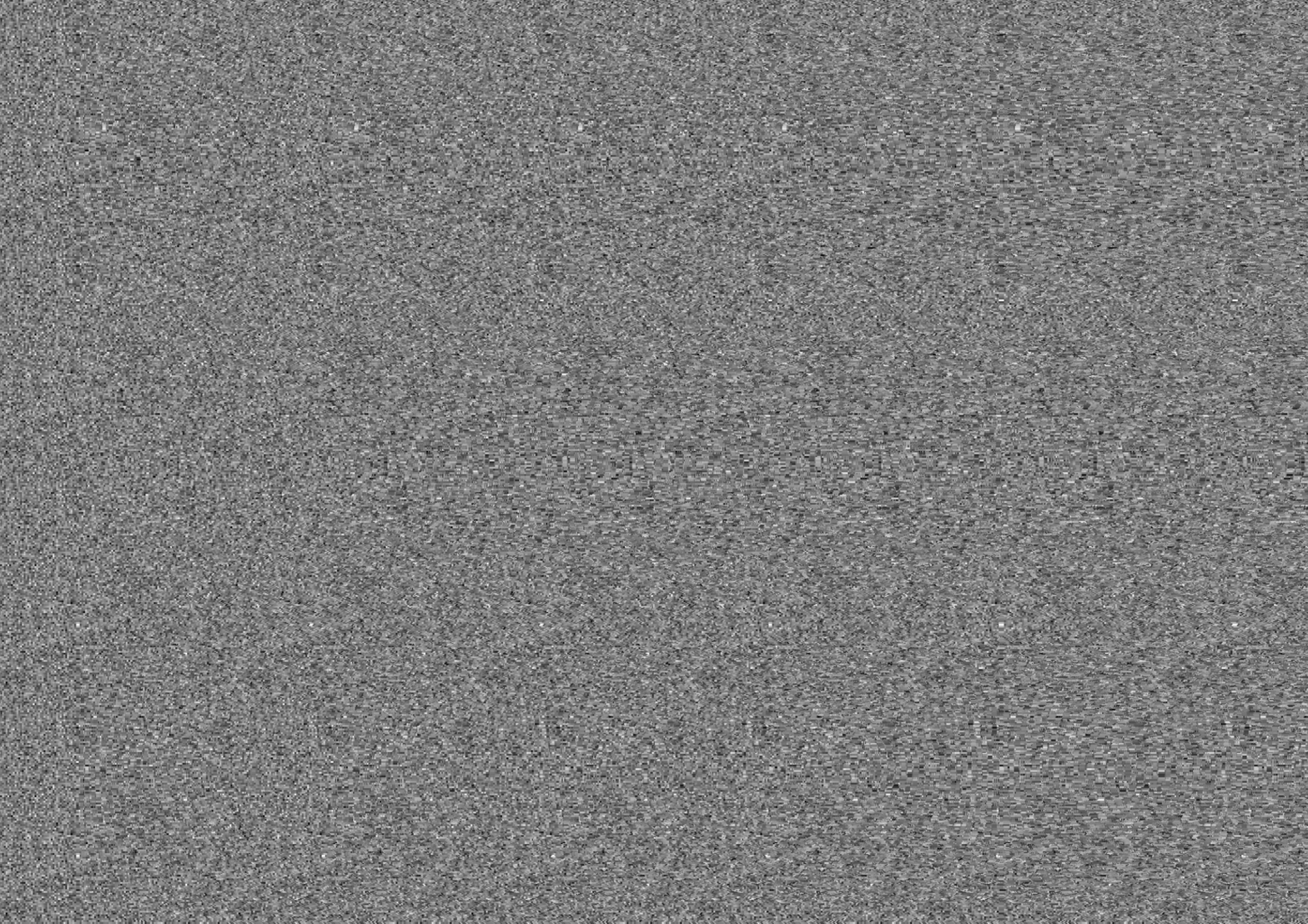}
\label{Fig:as_white}
}
\subfloat[]{
\centering \includegraphics[scale=0.23,clip,trim=0cm 0cm 0cm 0cm]{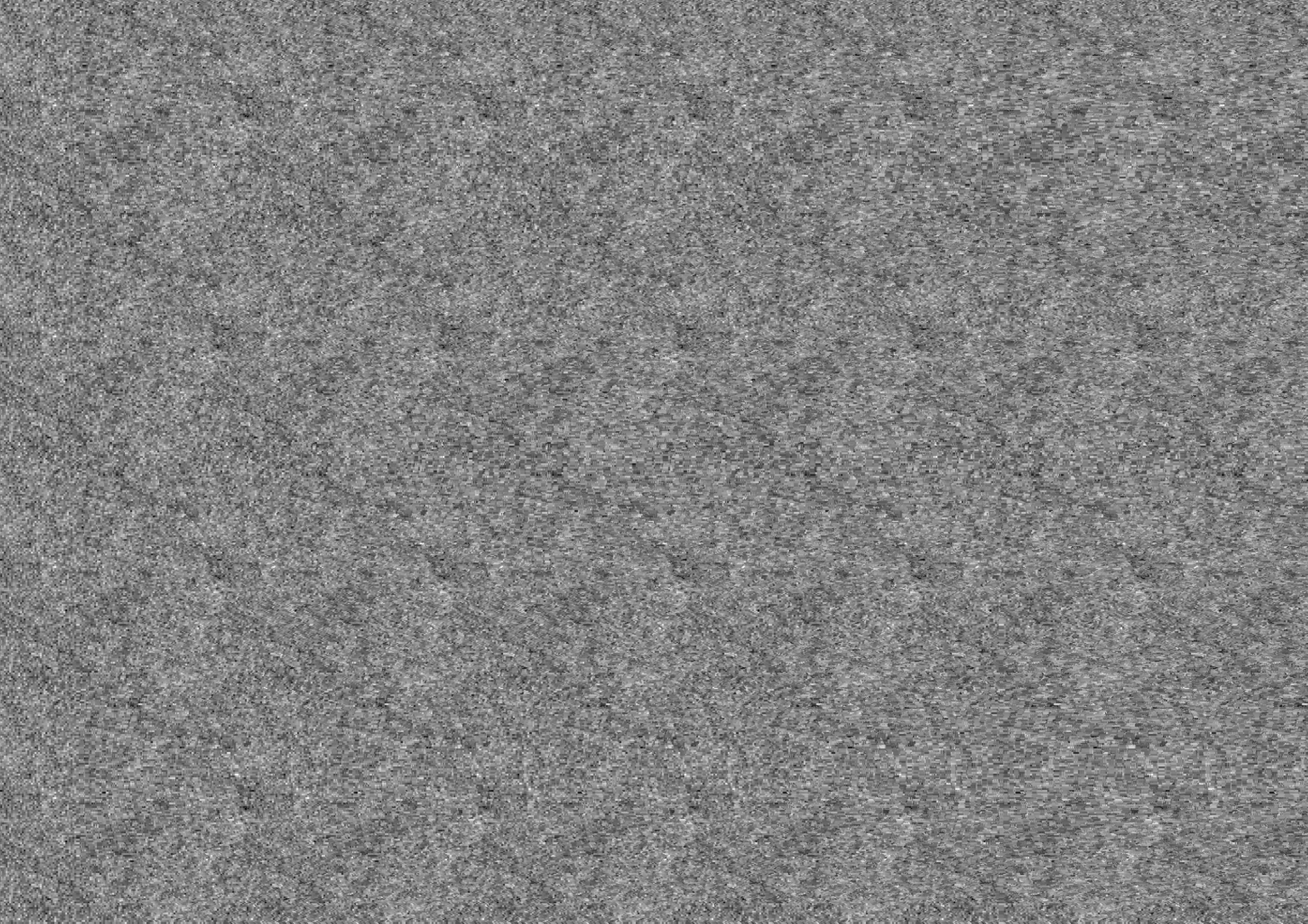}
\label{Fig:as_1_2}
}\\
\subfloat[]{
\centering \includegraphics[scale=0.23,clip,trim=0cm 0cm 0cm 0cm]{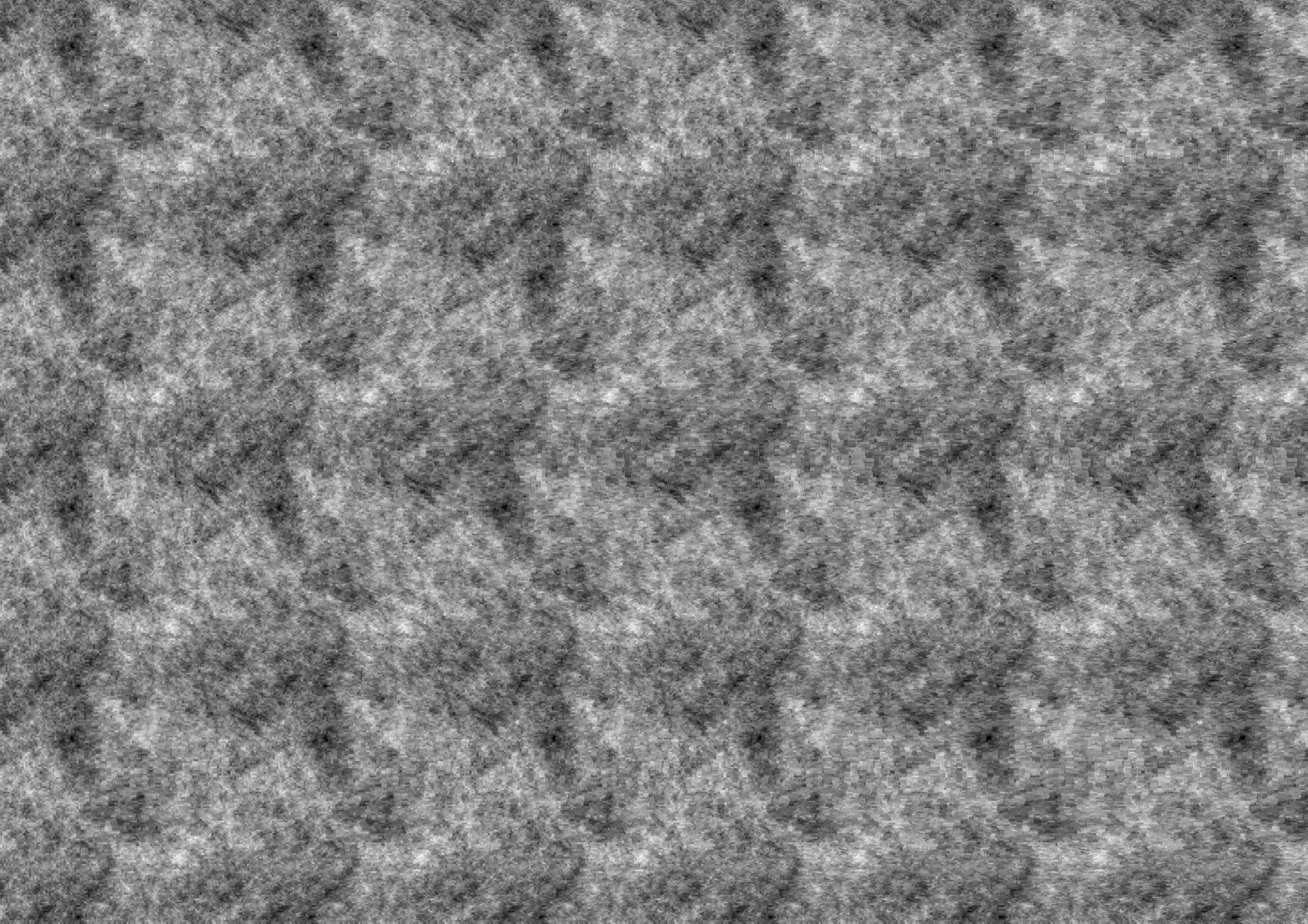}
\label{Fig:as_pink}
}
\centering
\subfloat[]{
\centering \includegraphics[scale=0.23,clip,trim=0cm 0cm 0cm 0cm]{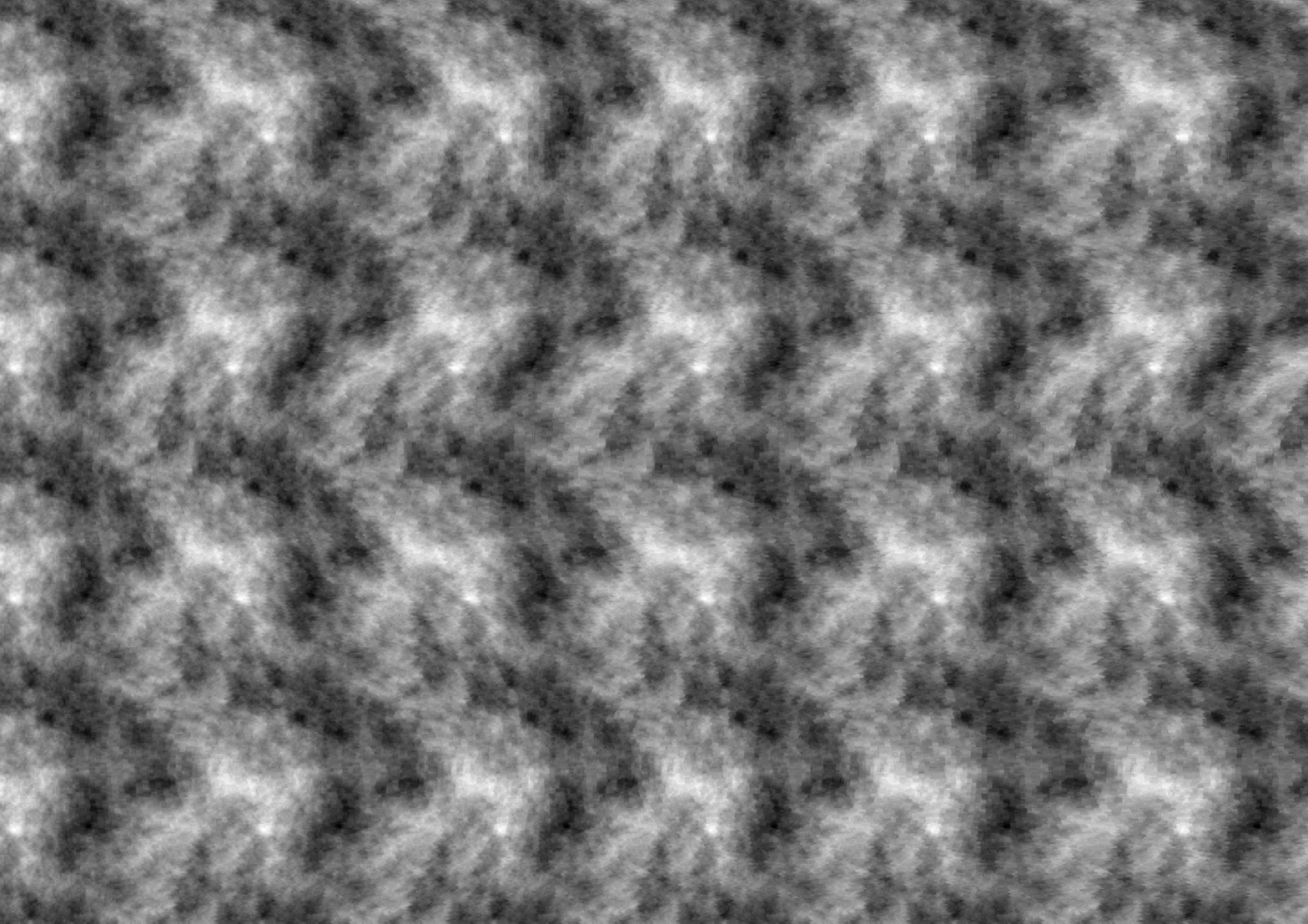}
\label{Fig:as_3_2}
}\\
\subfloat[]{
\centering \includegraphics[scale=0.23,clip,trim=0cm 0cm 0cm 0cm]{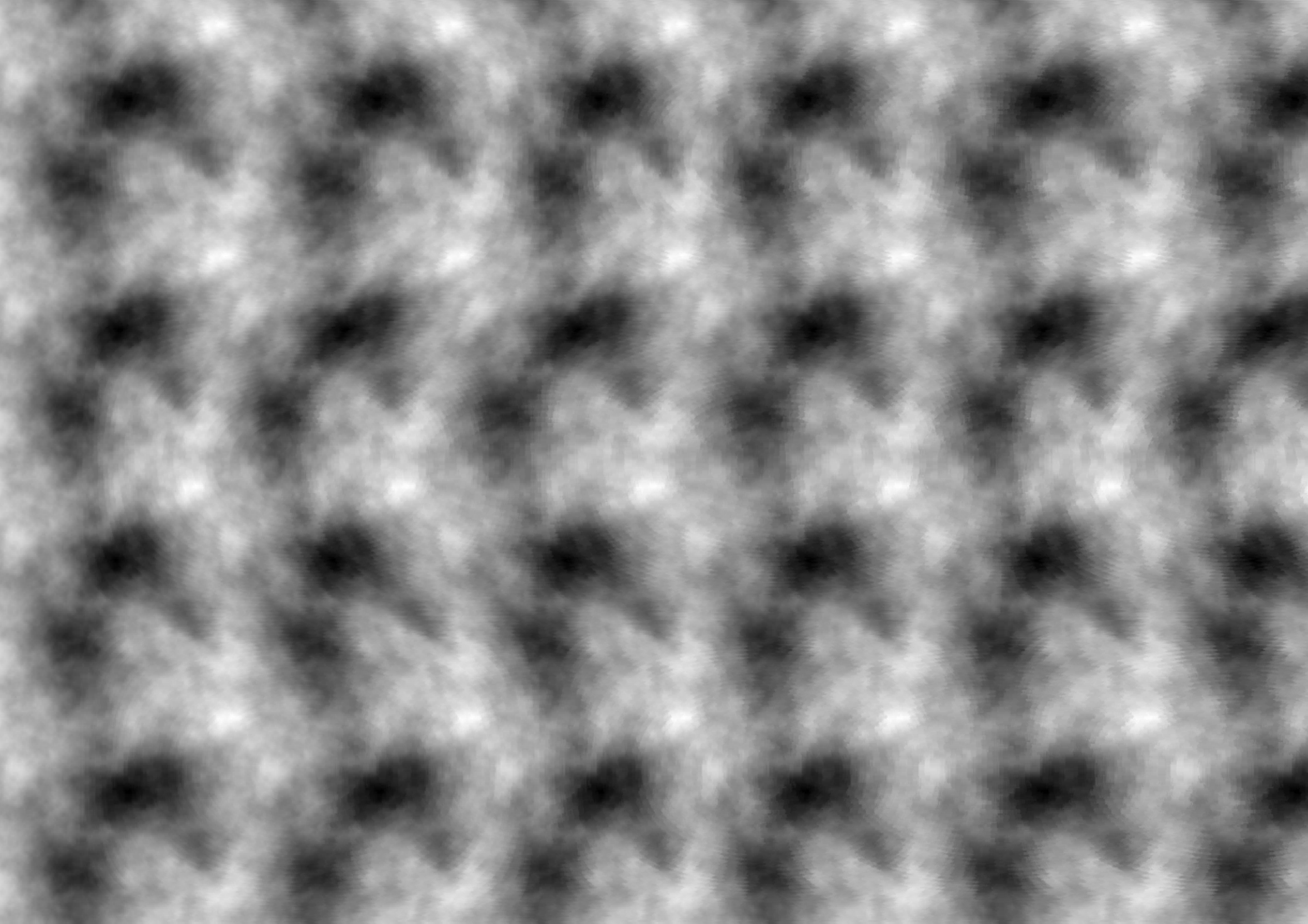}
\label{Fig:as_brown}
}

\caption[]{Examples of autostereograms of the egg crate depth map (Figure~\ref{Fig:eggcrate_depthmap}) with different noise patterns of the form $1/f^\beta$: \subref{Fig:as_white} $\beta=0$ (white noise), \subref{Fig:as_1_2} $\beta=\frac{1}{2}$, \subref{Fig:as_pink} $\beta=1$ (pink noise), \subref{Fig:as_3_2} $\beta=\frac{3}{2}$, \subref{Fig:as_brown} $\beta=2$ (brown noise). }
\label{Fig:AutostereogramsPerNoise}
\end{figure*}

The autostereograms were constructed using five noise patterns having a spectrum of $1/f^\beta$ with $\beta=0,\frac{1}{2},1,\frac{3}{2},2$.
Examples of the generated autostereograms for the different noise patterns are depicted in Figure~\ref{Fig:AutostereogramsPerNoise}.

A collection of 140 autostereograms was constructed for the experiment: each surface was used 7 times for each type of noise pattern. 

\subsubsection{Experiment 2 - Detail Discrimination}
High resolution detail in the depth dimension was created by using the depth profiles of four possible letters superimposed on the smooth surface of an ellipsoid (Figure~\ref{Fig:sphere_depthmap}).

Since the detail in depth rides on top of a smooth ellipsoidal surface, we in fact test the identification of higher resolution objects in the presence of a smooth, low resolution background.
With that regard, it should be noted that without the ellipsoid background, a ripple artifact appeared in the generated autostereograms, which could have led to the hidden letter being identified even without perceiving the depth dimension. 
\newline

There were four possible detail depth-profiles in the shape of the letters S, X, L, and T. In addition, there was a fifth option of no letter present, serving as a control option to check whether participants actually perceive the detail. 
The letters were $240\times 240$ pixels in size and were placed vertically in the middle of the surface with horizontal displacements of up to 400 pixels to the left or right. Those horizontal displacements were used so that participants wouldn't fixate on a single location in the autostereograms but instead search for the letter in the proximity of the center, while still placing the letters on a smooth background slope. 

The background surface was normalized to occupy 60\% of the gray level range, and the letters (when added) were scaled to 10\% of that range, i.e. higher than the highest point of the surface at a ratio of 1/6.
After incorporating the letters, a $5\times 5$ low-pass filter was applied to smooth the letter boundaries, in order to avoid miss-correspondence issues that could result in echoes in the generated autostereogram \cite{Thimbleby1994}.  
The letters were selected to be considerably different from one another in order to avoid confusion in their identification.

As in the first experiment, the autostereograms were generated using five noise patterns associated with $\beta=0,\frac{1}{2},1,\frac{3}{2},2$.
A collection of 125 autostereograms was constructed for the experiment: each letter, or lack thereof, was used 5 times in conjunction with each noise pattern. 

\begin{figure}[!htbp] 
\centering
\subfloat[]{
\centering \includegraphics[scale=0.13,clip,trim=0cm 0cm 0cm 0cm]{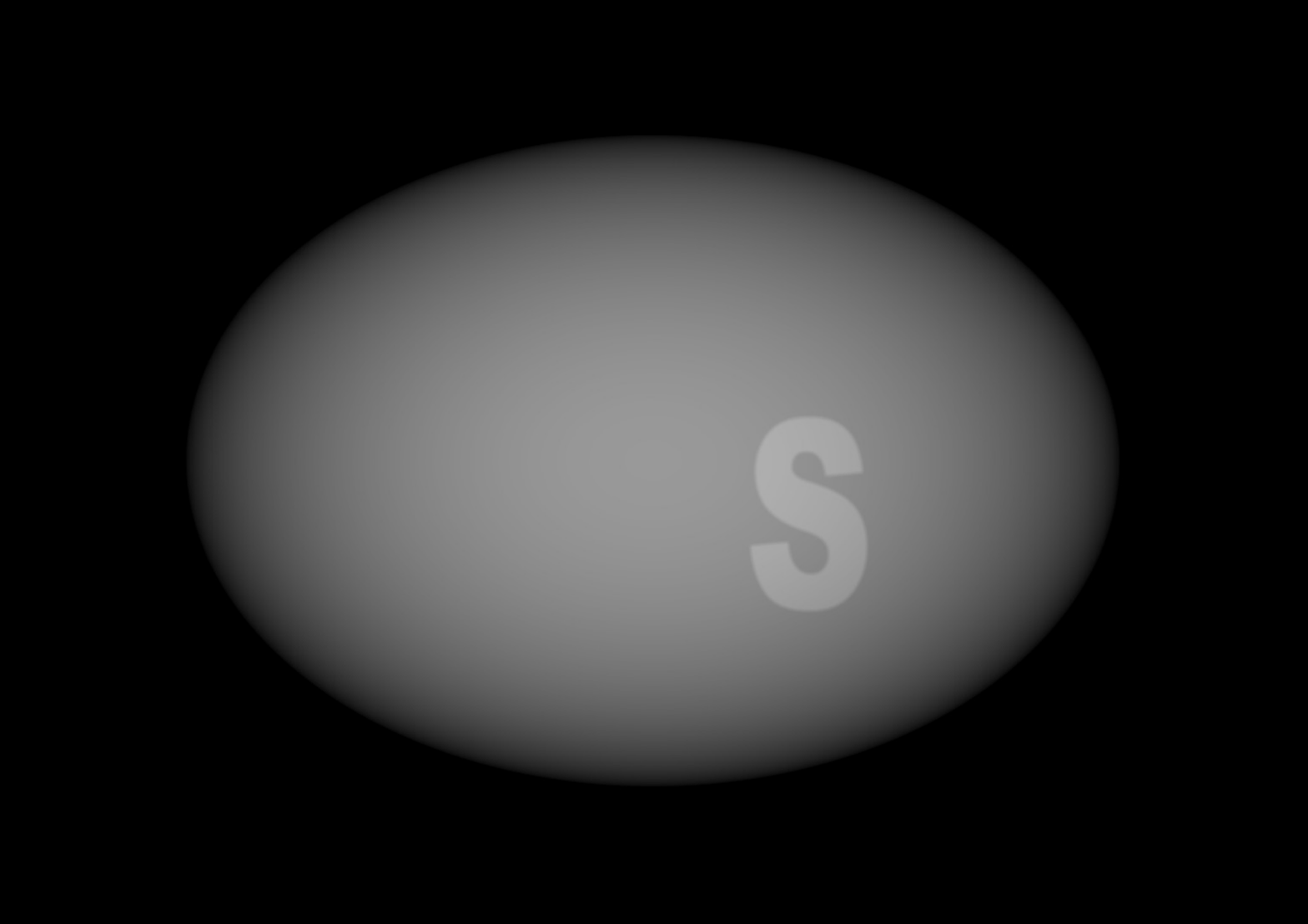}
\label{Fig:depthmapS}
}
\subfloat[]{
\centering \includegraphics[scale=0.13,clip,trim=0cm 0cm 0cm 0cm]{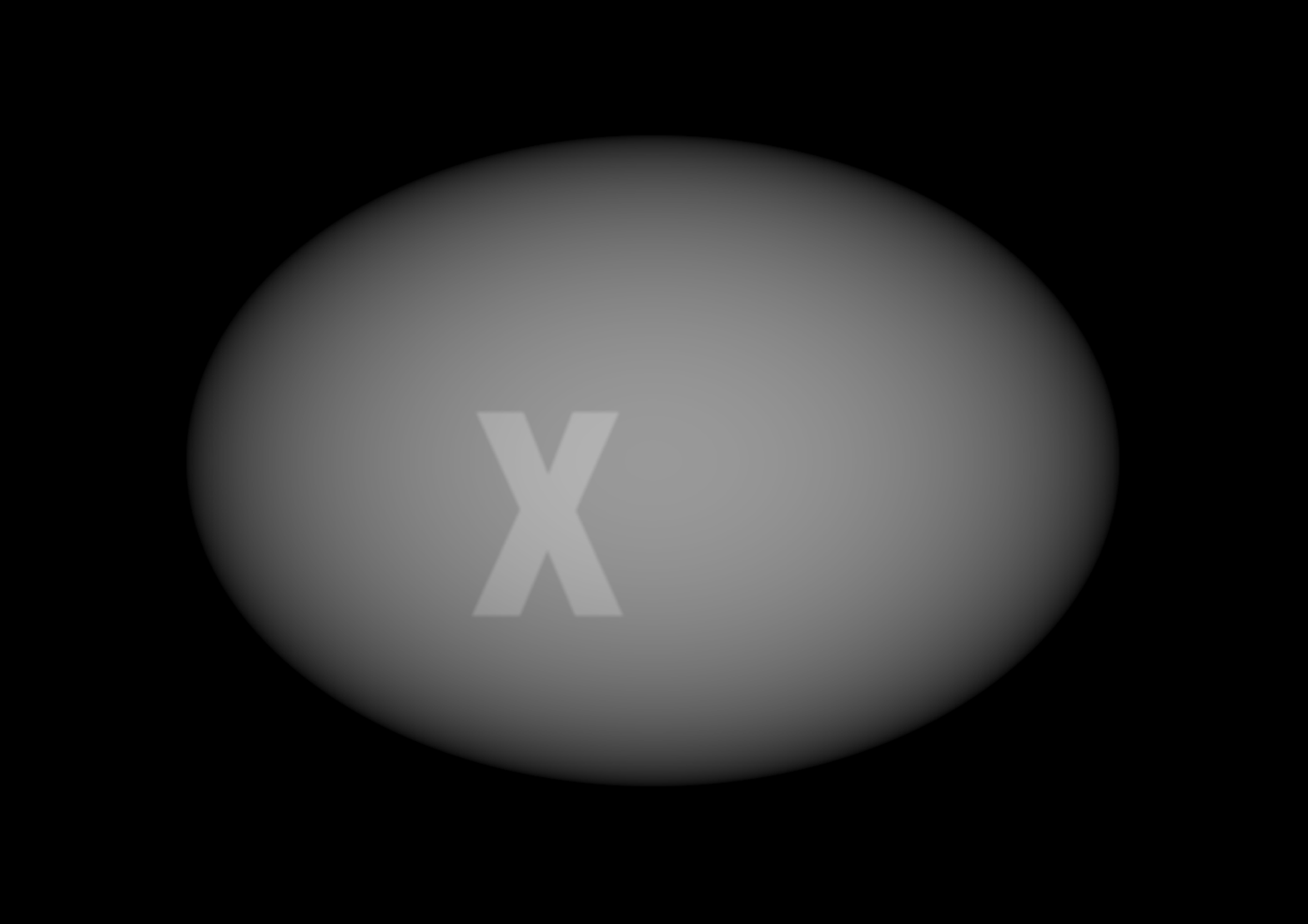}
\label{Fig:depthmapX}
}\\
\subfloat[]{
\centering \includegraphics[scale=0.13,clip,trim=0cm 0cm 0cm 0cm]{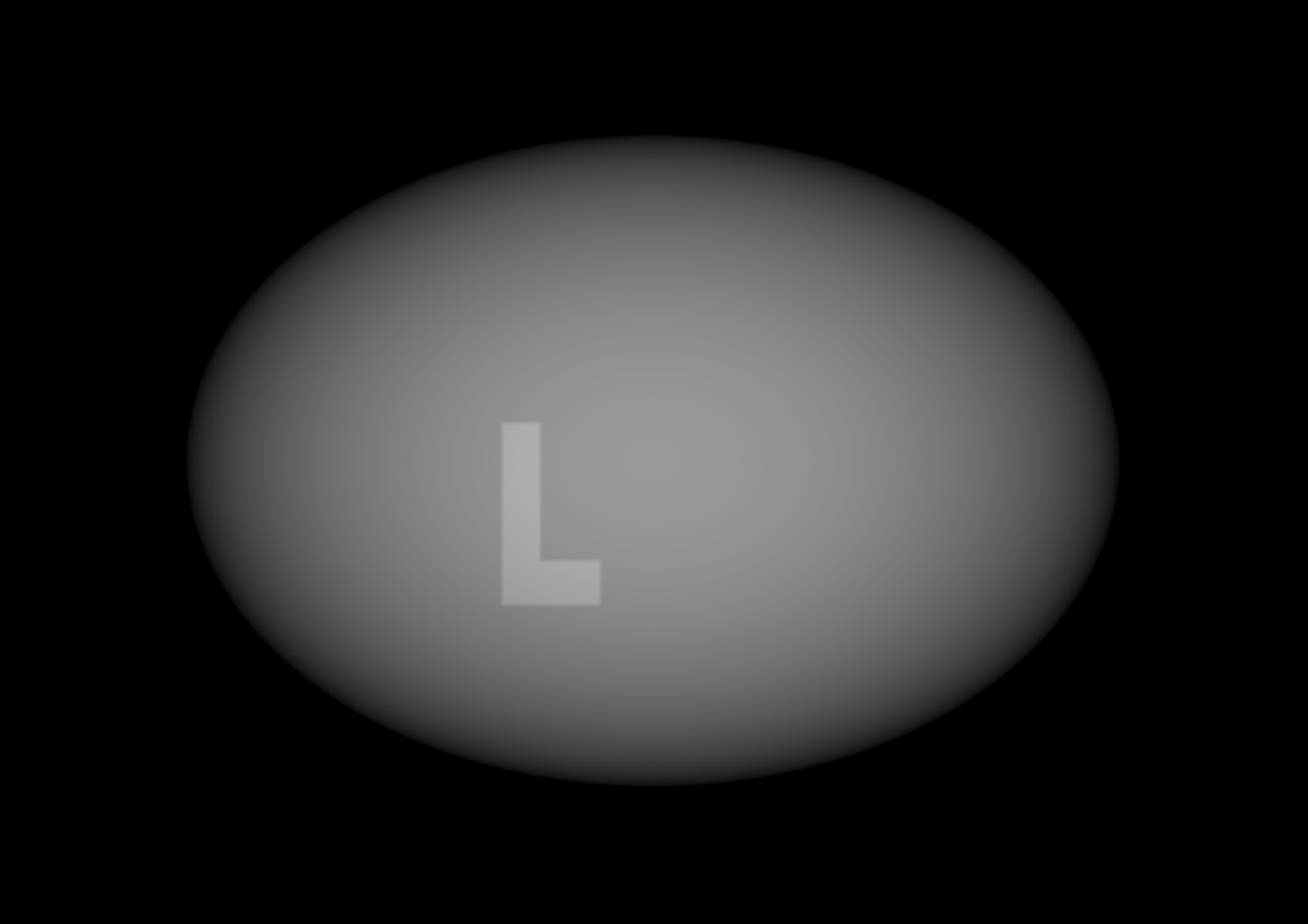}
\label{Fig:depthmapL}
} 
\subfloat[]{
\centering \includegraphics[scale=0.13,clip,trim=0cm 0cm 0cm 0cm]{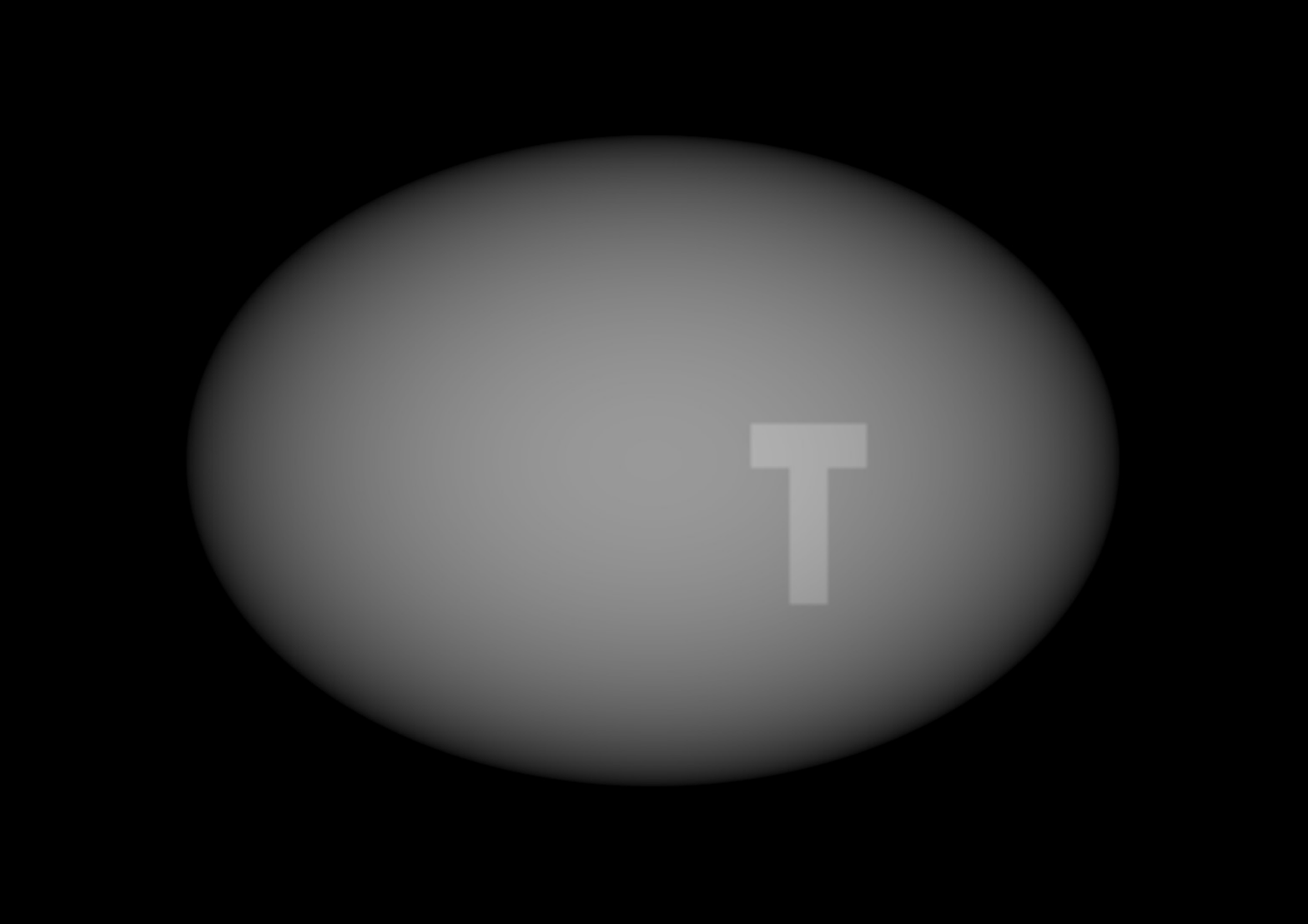}
\label{Fig:depthmapT}
}
\caption[]{Examples of depth maps used for experiment 2}
\label{Fig:DepthMaps_letters}
\end{figure}

Examples of depth maps with letters incorporated are presented in Figure~\ref{Fig:DepthMaps_letters}.

\subsubsection{Experiment 3 - Fine Detail Identification Limits}
Similar to the second experiment, high resolution in the depth dimension was represented by different letters superimposed on an ellipsoid surface (Figure~\ref{Fig:sphere_depthmap}).
The letters selected for this experiment were P and B. 
This time the letters were deliberately chosen to be not considerably different from one another in order to check for identification accuracy, and as opposed to the second experiment, there is always a letter inserted. 
In order to reach the limits of identification, the letter sizes were smaller: $20\times 20$, $40\times 40$, $60\times 60$, $80\times 80$, or $100\times 100$ pixels.

The background surface was again normalized to occupy 60\% of the gray level range, and the letters were scaled to 12\%,10\% or 8.57\% of that range, i.e. higher than the highest point of the surface at a ratio of 1/5,1/6 or 1/7.
After incorporating the letters, here too a $5\times 5$ low-pass filter was applied to smooth the resulting surface. 

The autostereograms were constructed using three noise patterns associated with $\beta=0,\frac{1}{2},1$.
A collection of 180 autostereograms was generated for the experiment: each letter was used twice for each combination of letter size and relative depth and for each noise pattern. 

\begin{figure}[!htbp] 
\centering
\subfloat[]{
\centering \includegraphics[scale=0.15,clip,trim=2cm 0cm 2cm 0cm]{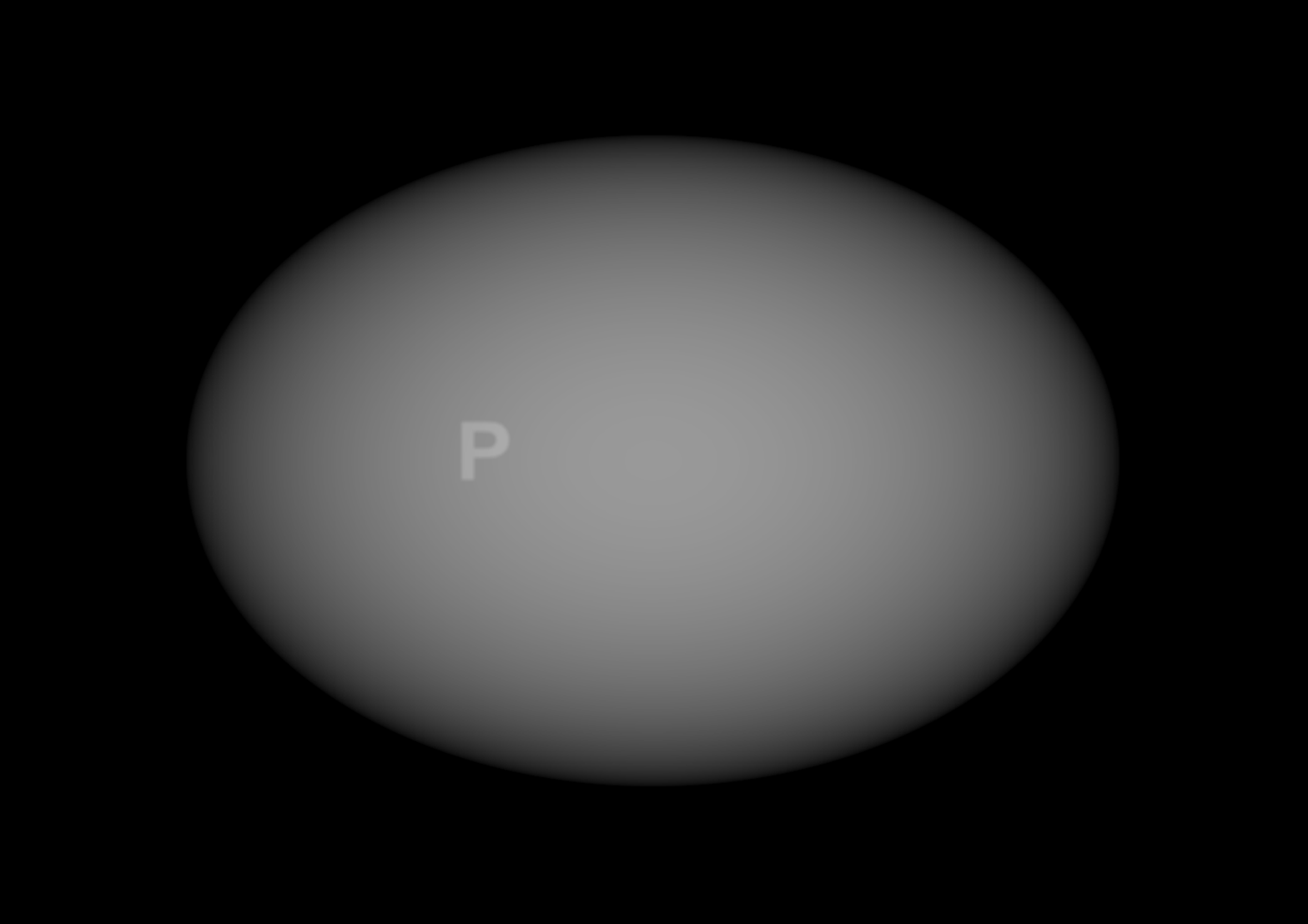}
\label{Fig:depthmapP}
}
\subfloat[]{
\centering \includegraphics[scale=0.15,clip,trim=2cm 0cm 2cm 0cm]{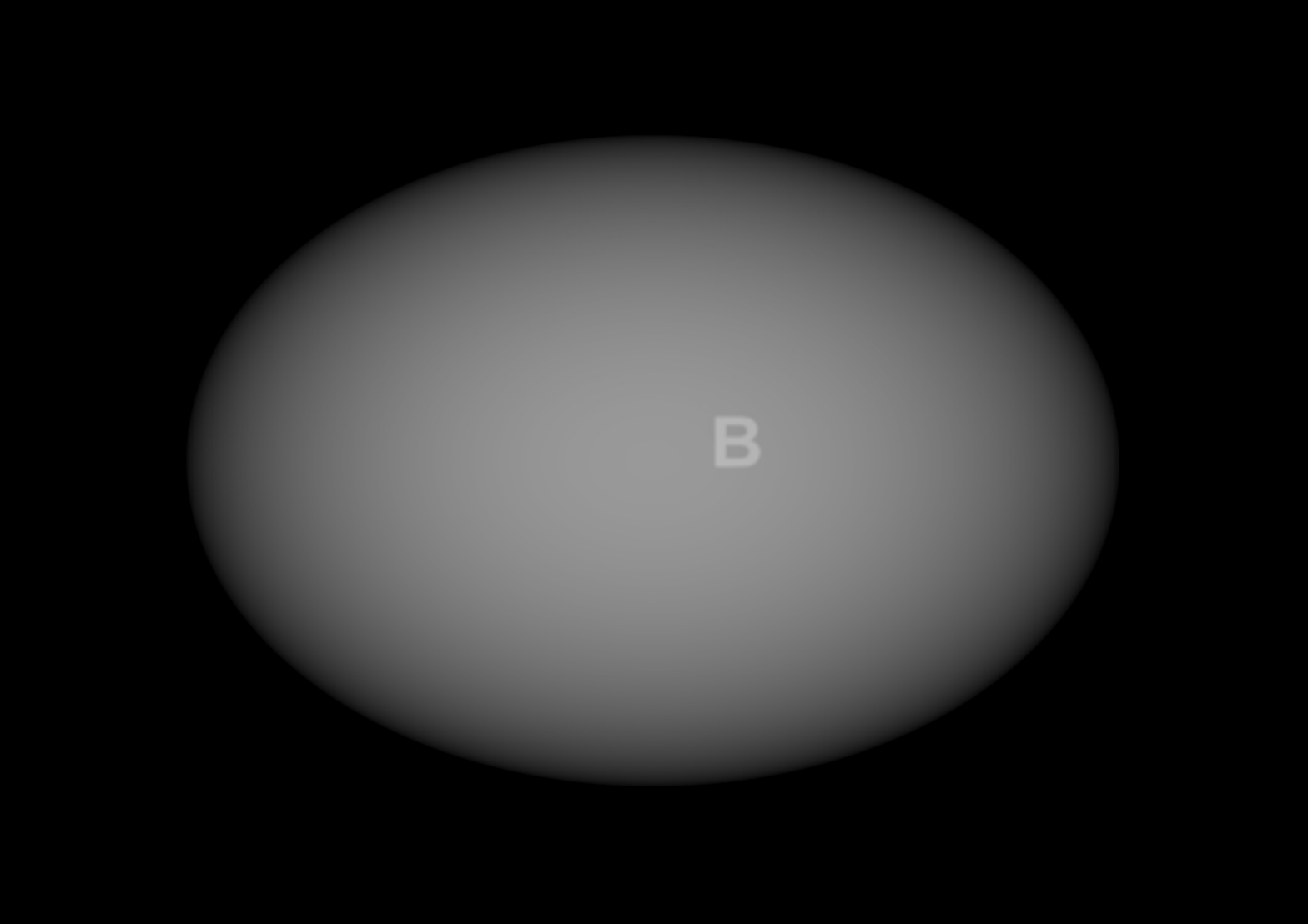}
\label{Fig:depthmapB}
}
\caption[]{Examples of depth maps used for experiment 3: \subref{Fig:depthmapP} The letter P ($100\times 100$ pixels),  \subref{Fig:depthmapB} The letter B ($80\times 80$ pixels)}
\label{Fig:DepthMaps_letters_jnd}
\end{figure}

Examples of the depth maps with incorporated letters used for the identification limit experiment are presented in Figure~\ref{Fig:DepthMaps_letters_jnd}.

\subsection{Procedure}

The experiments were conducted in an isolated room at the Intelligent Systems Lab (ISL) at the Technion. All the participants used the same computer and screen, with the same room illumination, and were shown the same autostereograms (in randomized order).
The autostereograms were displayed in full-screen mode on a $22"$ LCD monitor with a $1680\times 1050$ resolution.

The Psychophysics Toolbox Version 3 for MATLAB 
\cite{Brainard1997,Kleiner2007} was used for displaying the autostereograms and for collecting the results.

The procedure was as follows: 
An autostereogram was picked from the randomly ordered collection and displayed on the screen. 
The participants, their hand on the computer mouse at all time, had to press the left mouse button when perceiving the hidden surface.
A selection screen then appeared offering a choice between all the possible depth maps (surfaces or letters) and an ``undefinable" option.
After selecting an answer, a new autostereogram would be displayed and so forth until completing the set. 

\begin{figure}[!htbp] 
\centering
\subfloat[Experiment 1]{
\centering \fbox{\includegraphics[scale=0.16,clip,trim=1.8cm 7cm 20cm 2cm]{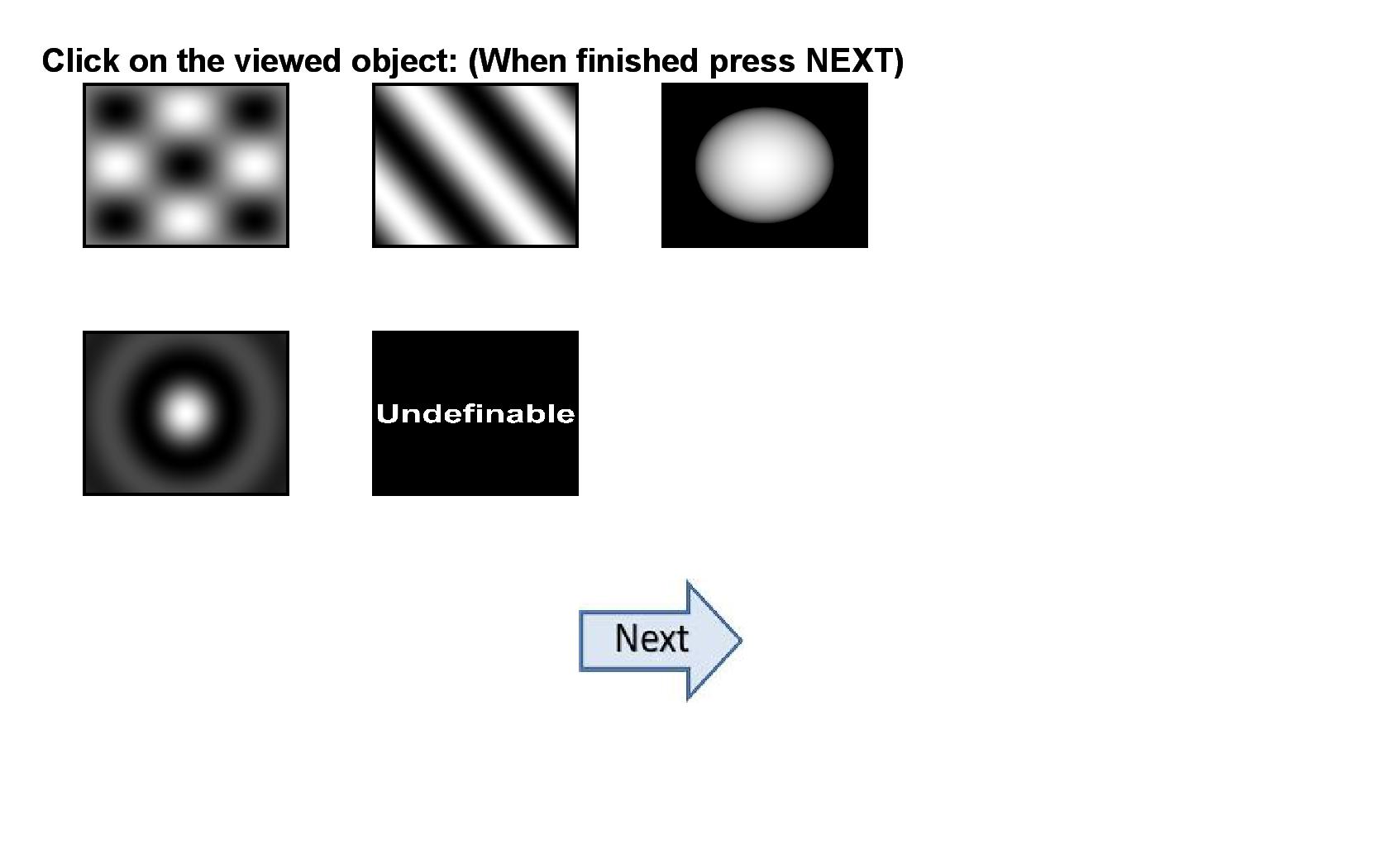}}
\label{Fig:selection0}
}\\
\subfloat[Experiment 2]{
\centering \fbox{\includegraphics[scale=0.16,clip,trim=1.8cm 7cm 20cm 2cm]{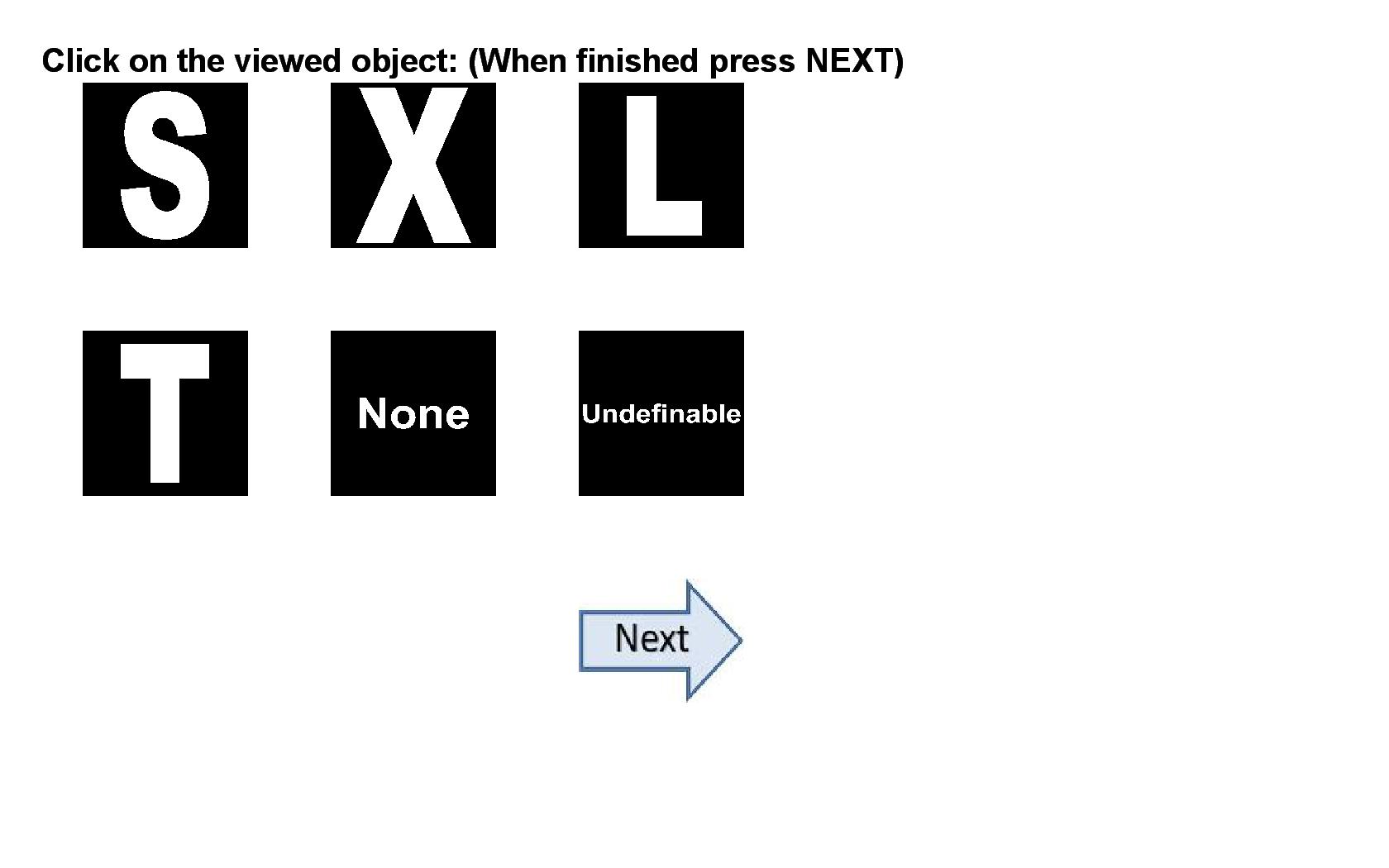}}
\label{Fig:selection1}
}\\
\subfloat[Experiment 3]{
\centering \fbox{\includegraphics[scale=0.16,clip,trim=1.8cm 7cm 20cm 2cm]{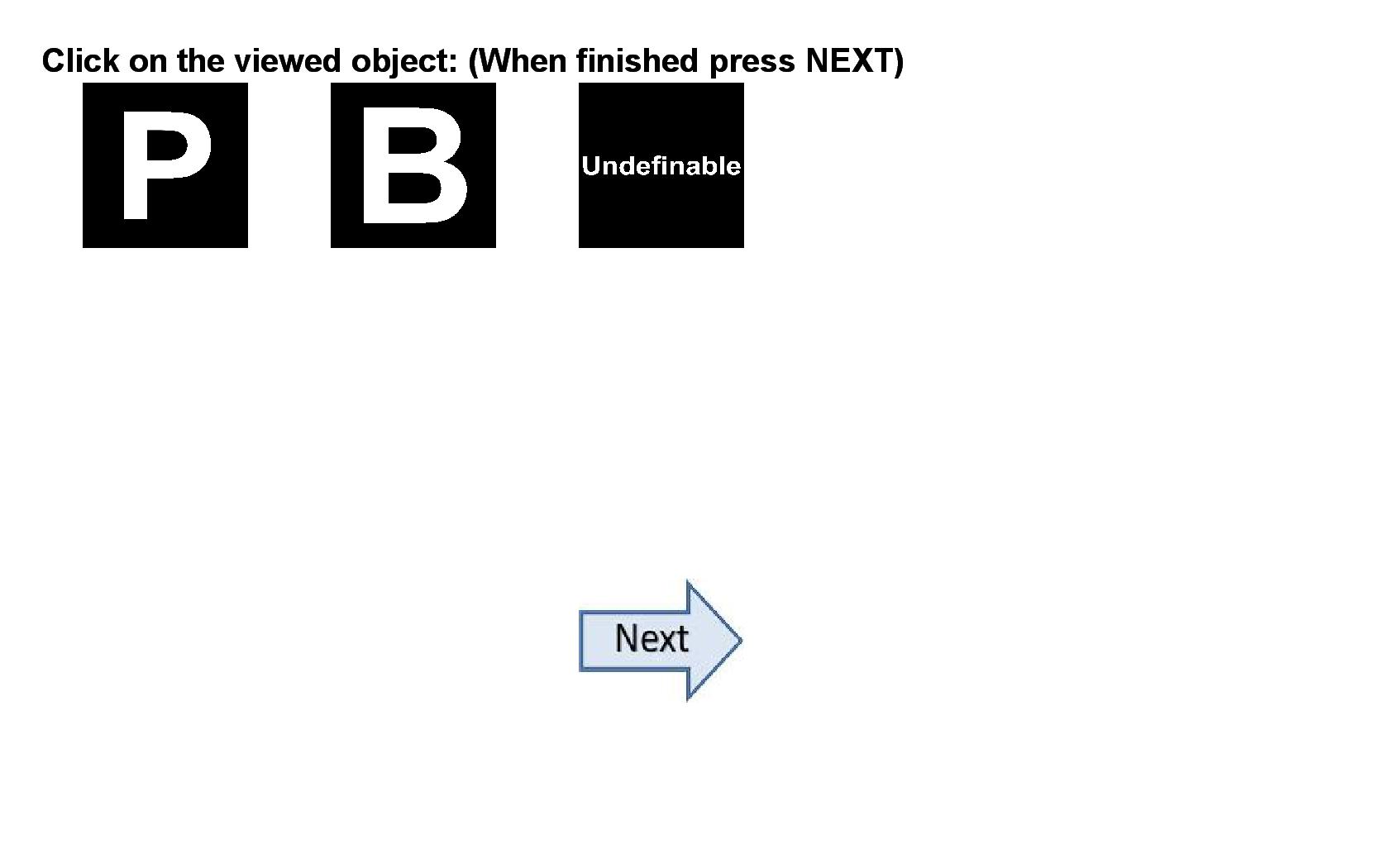}}
\label{Fig:selection2}
}
\caption[]{Selection screens used for the different experiments}
\label{Fig:SelectionScreens}
\end{figure}

The selection screen, besides collecting the results, resets the focus and convergence that were previously achieved by the participants \cite{Reimann1995}, thereby ensuring that the adjustment process starts from the same point each time.
The selection screens used for the different experiments are presented in Figure~\ref{Fig:SelectionScreens}.

The response time (RT) from the appearance of the autostereogram to the left button mouse click signaling identification was measured and the selection was checked for correctness. 

Since there was no time limit on the identification of the hidden object, participants were instructed to ``give up" after a long period of time and choose the ``undefinable" option in the selection screen. 

In each experiment, participants were first shown a training set of 10 autostereograms randomly chosen from the experiment's autostereograms pool so they will familiarize themselves with the test environment, calibrate their viewing position to their optimum and have a sense of when they need to ``give up". 
After the training set, a white screen appeared and the participants were instructed to press the left mouse button when ready to proceed to the actual test.

\section{Results and Discussion} \label{section:Results}
Participants were tested for accuracy and response time. 
Accuracy was measured by the percentage of correct answers and by the numbers of mistakes and choices of ``undefinable" for each noise. A mistake was counted when the selection was neither ``undefinable" nor the correct answer.
The mean and standard deviation (STD) of the measured response times were computed using only the correct responses.
The two samples one-tailed t-test was used to evaluate whether the mean response times obtained for different noise patterns have statistical significance.

\subsection{Experiment 1}
Table~\ref{tab:Exp1_correct15} shows the rate of correct answers and number of mistakes or choices of ``undefinable" versus the noise pattern used. 
All noise types exhibit high correctness rate, indicating that smooth depth maps are easily recognizable across all of the mentioned noise patterns. 
Yet a slightly lower accuracy can be observed for white noise, due to its relatively high amount of ``undefinable" selections.

Since no time limit was posed on identification, the accuracy is very high for all the noise patterns, and so we examined the differences in response times.

\begin{table*}[!htbp]
\centering
\scalebox{0.9}{
\begin{tabular}{|c||c|c|c|c|c|} 
\hline 
  & \begin{tabular}{c} White Noise \\ $\beta=0$ \end{tabular} & $\beta=\frac{1}{2}$ & \begin{tabular}{c} Pink Noise \\ $\beta=1$ \end{tabular} & $\beta=\frac{3}{2}$ & \begin{tabular}{c} Brown Noise \\ $\beta=2$ \end{tabular} \\
\hline \hline
Correct Rate & 99.05\% & 99.76\% & 99.52\% & 99.29\% & 99.52\% \\
\hline 
Number of Mistakes & 1 & 1 & 2 & 2 & 1 \\
\hline 
Number of Undefinables & 3 & 0 & 0 & 1 & 1 \\
\hline 
\end{tabular} 
}
\caption[]{Accuracy vs. Noise Patterns in the Surface Recognition Test (Experiment 1)} \label{tab:Exp1_correct15}
\end{table*}

\begin{figure}[htbp] 
\centering \includegraphics[scale=0.6,clip,trim=0.8cm 0cm 1.2cm 0cm]{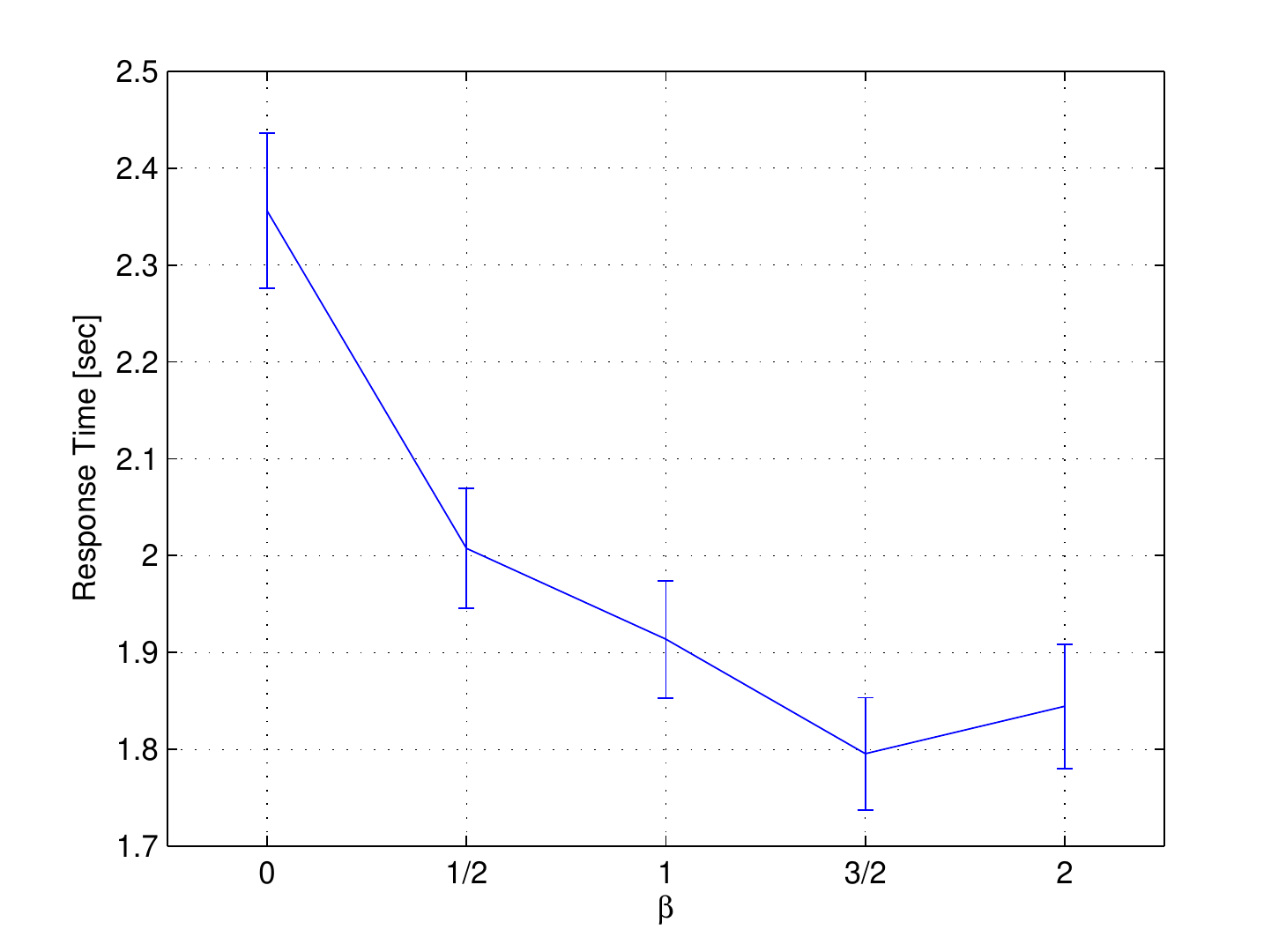} 
\caption[]{Mean Response Time vs. Noise Patterns in the Surface Recognition Test (Experiment 1). Error bars show standard error of the mean.}
\label{Fig:Exp1_IT15_outliers}
\end{figure}

Figure~\ref{Fig:Exp1_IT15_outliers} presents the mean response time (RT) versus noise pattern. 
While the mean RT is typically around 2 seconds, in 8 of the 2100 samples (which constitute 0.38\%) RTs over 10 seconds and upto 76 seconds were measured. Therefore we consider those samples as outliers and exclude them from our analysis.

A one-tailed t-test was performed on the response times of every pair of noise patterns to test whether the mean response times have statistical significance. 
The results of the t-test are presented in Table~\ref{tab:Exp1_ttest_outliers}.

\begin{table}[htbp] 
\centering
\begin{tabular}{|c||c|c|} 
\hline 
  Hypothesis & \begin{tabular}{c} Significant \\ (P-value$<$0.05) \end{tabular} & P-value \\
\hline \hline
$RT_{\beta=0}>RT_{\beta=\frac{1}{2}}$ & yes & 2.954e-4 \\
\hline 
$RT_{\beta=0}>RT_{\beta=1}$ & yes & 5.68e-6 \\
\hline 
$RT_{\beta=0}>RT_{\beta=\frac{3}{2}}$ & yes & 9.858e-9 \\
\hline 
$RT_{\beta=0}>RT_{\beta=2}$ & yes & 3.656e-7 \\
\hline 
$RT_{\beta=\frac{1}{2}}>RT_{\beta=1}$ & no & 0.1389 \\
\hline 
$RT_{\beta=\frac{1}{2}}>RT_{\beta=\frac{3}{2}}$ & yes & 0.0063 \\
\hline 
$RT_{\beta=\frac{1}{2}}>RT_{\beta=2}$ & yes & 0.0338 \\
\hline 
$RT_{\beta=1}>RT_{\beta=\frac{3}{2}}$ & no & 0.0794 \\
\hline 
$RT_{\beta=1}>RT_{\beta=2}$ & no & 0.216 \\
\hline 
$RT_{\beta=\frac{3}{2}}<RT_{\beta=2}$ & no & 0.2862 \\
\hline 
\end{tabular} 
\caption[]{T-Test Results for the Surface Recognition Test (Experiment 1)} \label{tab:Exp1_ttest_outliers}
\end{table}

It can be observed that in accordance with \cite{Bruckstein1996}, smooth surfaces hidden in autostereograms made of white noise patterns are significantly harder to perceive than with any other noise pattern. 
The best results are obtained for $\beta=\frac{3}{2}$, with non-significant difference from performance with $1\leq\beta\leq 2$.

\subsection{Experiment 2}
Table~\ref{tab:Exp2_correct} shows the rate of correct answers and number of mistakes or choices of ``undefinable" versus the noise pattern used in experiment 2. 
All the noise patterns exhibit high accuracy in letter recognition, with a slight deterioration observed for brown noise ($\beta =2$). 

\begin{table*}[!htbp]
\centering
\scalebox{0.9}{
\begin{tabular}{|c||c|c|c|c|c|} 
\hline 
  & \begin{tabular}{c} White Noise \\ $\beta=0$ \end{tabular} & $\beta=\frac{1}{2}$ & \begin{tabular}{c} Pink Noise \\ $\beta=1$ \end{tabular} & $\beta=\frac{3}{2}$ & \begin{tabular}{c} Brown Noise \\ $\beta=2$ \end{tabular} \\
\hline \hline
Correct Rate & 100\% & 99.47\% & 100\% & 99.47\% & 97.07\% \\
\hline 
Number of Mistakes & 0 & 0 & 0 & 0 & 6 \\
\hline 
Number of Undefinables & 0 & 2 & 0 & 2 & 5 \\
\hline 
\end{tabular} 
}
\caption[]{Accuracy vs. Noise Patterns in the Detail Discrimination Test (Experiment 2)} \label{tab:Exp2_correct}
\end{table*}

\begin{figure}[h] 
\centering \includegraphics[scale=0.6,clip,trim=0.8cm 0cm 1.2cm 0cm]{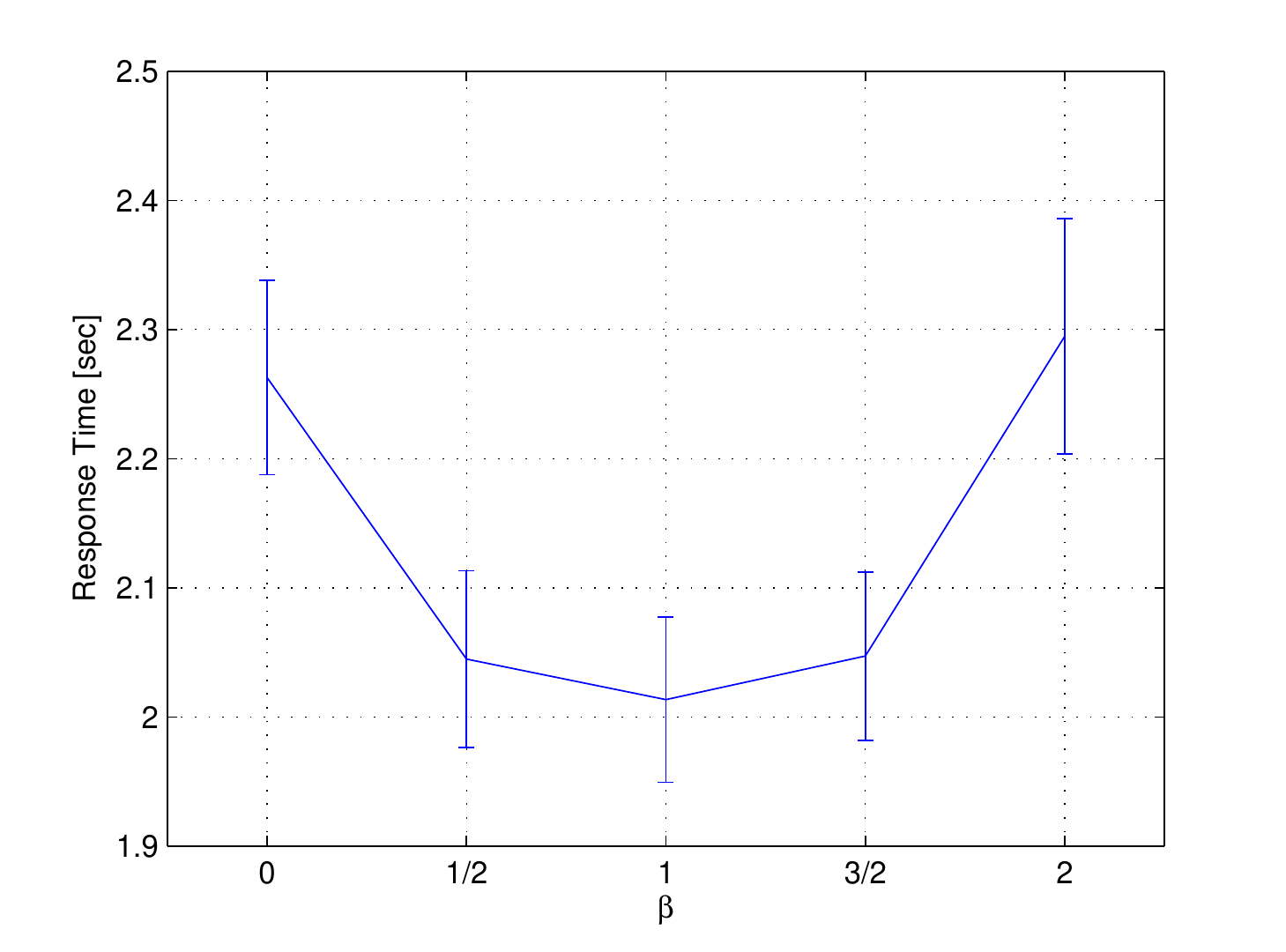}
\caption[]{Mean Response Time vs. Noise Patterns in the Detail Discrimination Test (Experiment 2)}
\label{Fig:Exp2_IT}
\end{figure}

Figure~\ref{Fig:Exp2_IT} presents the mean and standard error of the response time versus noise pattern.
T-test results are presented in Table~\ref{tab:Exp2_ttest}.

\begin{table}[h]
\centering
\begin{tabular}{|c||c|c|} 
\hline 
  Hypothesis & \begin{tabular}{c} Significant \\ (P-value$<$0.05) \end{tabular} & P-value \\
\hline \hline
$RT_{\beta=0}>RT_{\beta=\frac{1}{2}}$ & yes & 0.0162 \\
\hline 
$RT_{\beta=0}>RT_{\beta=1}$ & yes & 0.0059 \\
\hline 
$RT_{\beta=0}>RT_{\beta=\frac{3}{2}}$ & yes & 0.0152 \\
\hline 
$RT_{\beta=0}<RT_{\beta=2}$ & no & 0.3938 \\
\hline 
$RT_{\beta=\frac{1}{2}}>RT_{\beta=1}$ & no & 0.3689 \\
\hline 
$RT_{\beta=\frac{1}{2}}<RT_{\beta=\frac{3}{2}}$ & no & 0.4902 \\
\hline 
$RT_{\beta=\frac{1}{2}}<RT_{\beta=2}$ & yes & 0.0144 \\
\hline 
$RT_{\beta=1}<RT_{\beta=\frac{3}{2}}$ & no & 0.356 \\
\hline 
$RT_{\beta=1}<RT_{\beta=2}$ & yes & 0.0059 \\
\hline 
$RT_{\beta=\frac{3}{2}}<RT_{\beta=2}$ & yes & 0.0137 \\
\hline 
\end{tabular} 
\caption[]{T-Test Results for the Detail Discrimination Test (Experiment 2)} \label{tab:Exp2_ttest}
\end{table}

It can be observed that the best performance is obtained for noise spectra with $\frac{1}{2}\leq\beta\leq\frac{3}{2}$ with non-significant difference between them.
Brown noise is inferior to all other noise patterns in terms of both accuracy and response time.
White noise exhibits high response time, presumably due to the amount of time it takes to first identify the smooth background (experiment 1), but high correctness rate with no mistakes made by the participants.

\subsection{Experiment 3}
The results of the previous experiment indicate that pink noise is superior to white noise in terms of response time in identification of high resolution details. 
However, in order to assure that this didn't result from the incorporated letters being too big, the third experiment compares these noises again for significantly smaller letters, where white noise is expected to have an advantage over the other noise patterns. Brown noise ($\beta=2$) wasn't included in this experiment as it is expected to perform worst for fine detail identification.

Since our aim in this experiment was testing perception limits, identification was not limited in time and we focused our analysis mainly on the identification accuracy.

\begin{figure}[!htbp] 
\centering \includegraphics[scale=0.6,clip,trim=0.8cm 0cm 1.2cm 0.7cm]{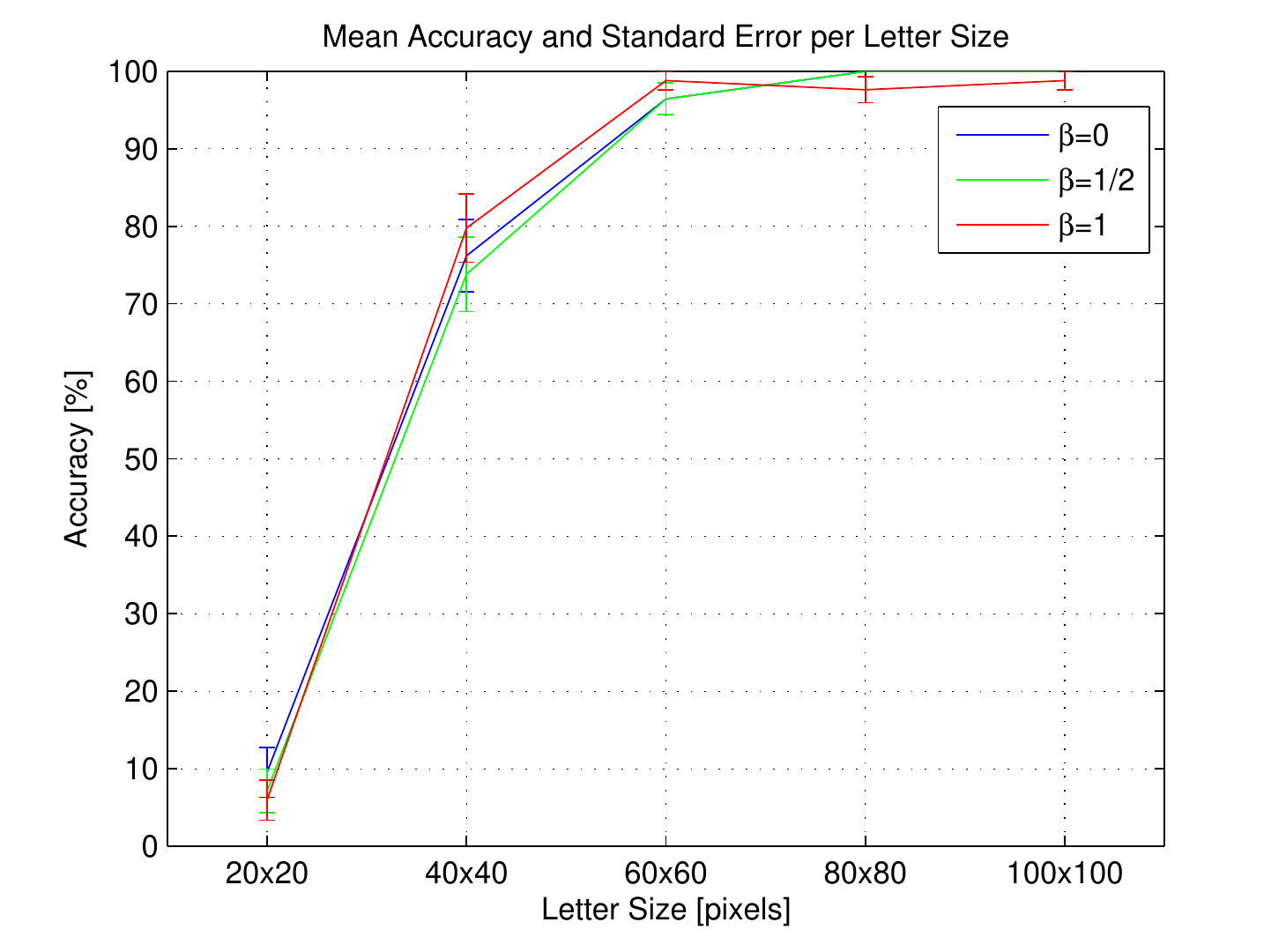}
\caption[]{Correct Rate vs. Letter Size in the identification limit test (Experiment 3)}
\label{Fig:Exp3_correct}
\end{figure}

The correct rate as a function of the letter size is presented in Figure~\ref{Fig:Exp3_correct}. 
Data was accumulated and averaged for 3 different relative depths of the letters with respect to the background map. 

Table~\ref{tab:Exp3_table_noises} displays the percentage of correct answers, mistakes and ``undefinable" choices per noise, calculated for all participants as a function of both letter size and relative depth.

\begin{table*}[!htb]
\centering
\scalebox{0.8}{
\small
\begin{tabu} to \textwidth {|>{\bfseries}c||c|c|>{\bfseries}c||c|c|>{\bfseries}c||c|c|>{\bfseries}c||c|c|>{\bfseries}c||c|c|>{\bfseries}c||c|c|>{\bfseries}c|} 

\hline
\rowfont{\bfseries}
\multicolumn{1}{|c||}{\multirow{2}{*}{Depth$\backslash$Size} } 
& \multicolumn{3}{c||}{$\bold{20\times 20}$} & \multicolumn{3}{c||}{$\bold{40\times 40}$} & \multicolumn{3}{c||}{$\bold{60\times 60}$} & \multicolumn{3}{c||}{$\bold{80\times 80}$} & \multicolumn{3}{c||}{$\bold{100\times 100}$} & \multicolumn{3}{c|}{Row Mean}\\
\cline{2-19}
\rowfont{\bfseries}
\multicolumn{1}{|c||}{} & u & m & c & u & m & c & u & m & c & u & m & c & u & m & c & u & m & c \\
\hline \hline
1/5 & 85.71 & 0 &  \textcolor{cyan}{14.29} & 14.29 & 3.57 & \textcolor{cyan}{82.14} & 3.57 & 0 & \textcolor{cyan}{96.43} & 0 & 0 & \textcolor{cyan}{100} & 0 & 0 & \textcolor{cyan}{100} & 20.71 & 0.71 & \textcolor{blue}{78.57} \\
\hline 
1/6 & 82.14 & 10.7 & \textcolor{cyan}{7.14} & 32.14 & 3.57 & \textcolor{cyan}{64.29} & 3.57 & 0 & \textcolor{cyan}{96.43} & 0 & 0 & \textcolor{cyan}{100} & 0 & 0 & \textcolor{cyan}{100} & 23.57 & 2.86 & \textcolor{blue}{73.57} \\
\hline 
1/7 & 92.86 & 0 & \textcolor{cyan}{7.14} & 14.29 & 3.57 & \textcolor{cyan}{82.14} & 3.57 & 0 & \textcolor{cyan}{96.43} & 0 & 0 & \textcolor{cyan}{100} & 0 & 0 & \textcolor{cyan}{100} & 22.14 & 0.71 & \textcolor{blue}{77.14} \\
\hline \hline
Col. Mean & 86.9 & 3.57 & \textcolor{blue}{9.52} & 20.24 & 3.57 & \textcolor{blue}{76.19} & 3.57 & 0 & \textcolor{blue}{96.43} & 0 & 0 & \textcolor{blue}{100} & 0 & 0 & \textcolor{blue}{100} & \multicolumn{3}{c}{}  \\
\cline{1-16}
\multicolumn{19}{c}{} \\
\multicolumn{19}{c}{\normalsize $\boldsymbol{\beta=0}$} \\
\multicolumn{19}{c}{} \\
\hline
\rowfont{\bfseries}
\multicolumn{1}{|c||}{\multirow{2}{*}{Depth$\backslash$Size} } 
& \multicolumn{3}{c||}{$\bold{20\times 20}$} & \multicolumn{3}{c||}{$\bold{40\times 40}$} & \multicolumn{3}{c||}{$\bold{60\times 60}$} & \multicolumn{3}{c||}{$\bold{80\times 80}$} & \multicolumn{3}{c||}{$\bold{100\times 100}$} & \multicolumn{3}{c|}{Row Mean}\\
\cline{2-19}
\rowfont{\bfseries}
\multicolumn{1}{|c||}{} & u & m & c & u & m & c & u & m & c & u & m & c & u & m & c & u & m & c \\
\hline \hline
1/5 & 89.29 & 7.14 & \textcolor{cyan}{3.57} & 17.89 & 0 & \textcolor{cyan}{82.14} & 0 & 3.57 & \textcolor{cyan}{96.43} & 0 & 0 & \textcolor{cyan}{100} & 0 & 0 & \textcolor{cyan}{100} & 21.43 & 2.14 & \textcolor{blue}{76.43} \\
\hline 
1/6 & 85.71 & 3.57 & \textcolor{cyan}{10.71} & 35.71 & 7.14 & \textcolor{cyan}{57.14} & 0 & 3.57 & \textcolor{cyan}{96.43} & 0 & 0 & \textcolor{cyan}{100} & 0 & 0 & \textcolor{cyan}{100} & 24.29 & 2.86 & \textcolor{blue}{72.86} \\
\hline 
1/7 & 89.29 & 3.57 & \textcolor{cyan}{7.14} & 17.86 & 0 & \textcolor{cyan}{82.14} & 3.57 & 0 & \textcolor{cyan}{96.43} & 0 & 0 & \textcolor{cyan}{100} & 0 & 0 & \textcolor{cyan}{100} & 22.14 & 0.71 & \textcolor{blue}{77.14} \\
\hline \hline
Col. Mean & 88.1 & 4.76 & \textcolor{blue}{7.14} & 23.81 & 2.38 & \textcolor{blue}{73.81} & 1.19 & 2.38 & \textcolor{blue}{96.43} & 0 & 0 & \textcolor{blue}{100} & 0 & 0 & \textcolor{blue}{100} & \multicolumn{3}{c}{}  \\
\cline{1-16}
\multicolumn{19}{c}{} \\
\multicolumn{19}{c}{\normalsize $\boldsymbol{\beta=\frac{1}{2}}$} \\
\multicolumn{19}{c}{} \\
\hline
\rowfont{\bfseries}
\multicolumn{1}{|c||}{\multirow{2}{*}{Depth$\backslash$Size} } 
& \multicolumn{3}{c||}{$\bold{20\times 20}$} & \multicolumn{3}{c||}{$\bold{40\times 40}$} & \multicolumn{3}{c||}{$\bold{60\times 60}$} & \multicolumn{3}{c||}{$\bold{80\times 80}$} & \multicolumn{3}{c||}{$\bold{100\times 100}$} & \multicolumn{3}{c|}{Row Mean}\\
\cline{2-19}
\rowfont{\bfseries}
\multicolumn{1}{|c||}{} & u & m & c & u & m & c & u & m & c & u & m & c & u & m & c & u & m & c \\
\hline \hline
1/5 & 85.71 & 7.14 &  \textcolor{cyan}{7.14} & 10.7 & 7.14 & \textcolor{cyan}{82.14} & 0 & 0 & \textcolor{cyan}{100} & 3.57 & 3.57 & \textcolor{cyan}{92.86} & 0 & 0 & \textcolor{cyan}{100} & 20 & 3.57 & \textcolor{blue}{76.43} \\
\hline 
1/6 & 96.43 & 0 & \textcolor{cyan}{3.57} & 35.7 & 0 & \textcolor{cyan}{64.29} & 3.57 & 0 & \textcolor{cyan}{96.43} & 0 & 0 & \textcolor{cyan}{100} & 0 & 0 & \textcolor{cyan}{100} & 27.14 & 0 & \textcolor{blue}{72.86} \\
\hline 
1/7 & 92.86 & 0 & \textcolor{cyan}{7.14} & 7.14 & 0 & \textcolor{cyan}{92.86} & 0 & 0 & \textcolor{cyan}{100} & 0 & 0 & \textcolor{cyan}{100} & 3.57 & 0 & \textcolor{cyan}{96.43} & 20.71 & 0 & \textcolor{blue}{79.29} \\
\hline \hline
Col. Mean & 91.67 & 2.38 & \textcolor{blue}{5.95} & 17.9 & 2.38 & \textcolor{blue}{79.76} & 1.19 & 0 & \textcolor{blue}{98.81} & 1.19 & 1.19 & \textcolor{blue}{97.62} & 1.19 & 0 & \textcolor{blue}{98.81} & \multicolumn{3}{c}{}  \\
\cline{1-16}
\multicolumn{19}{c}{} \\
\multicolumn{19}{c}{\normalsize $\boldsymbol{\beta=1}$} \\
\multicolumn{19}{c}{} \\
\end{tabu} 
}
\caption[]{Rate [\%] of Correct (c), Mistaken (m) and Undefinable (u) Selections per Letter Size and Relative Depth, for $\beta=0$ (White Noise), $\beta=\frac{1}{2}$ and $\beta=1$ (Pink Noise)} \label{tab:Exp3_table_noises}
\end{table*}

It can be observed that for all the tested noises, accuracy increases with letter size. 
However, its dependence on the relative depth seems quite random.
A previous experiment we performed, testing a wider range of relative depths, also didn't reveal any clear dependency.
The letter's size is therefore much more significant to its correct identification than its relative depth.  

On a one-tailed t-test we found no significance in accuracy between the different noise patterns. 
The lack of significance indicates that pink noise performs comparably well for the highest resolutions, as predicted by \cite{Bruckstein1996}.
\newline

\begin{figure}[!htbp] 
\centering \includegraphics[scale=0.6,clip,trim=0.8cm 0cm 1.2cm 0.7cm]{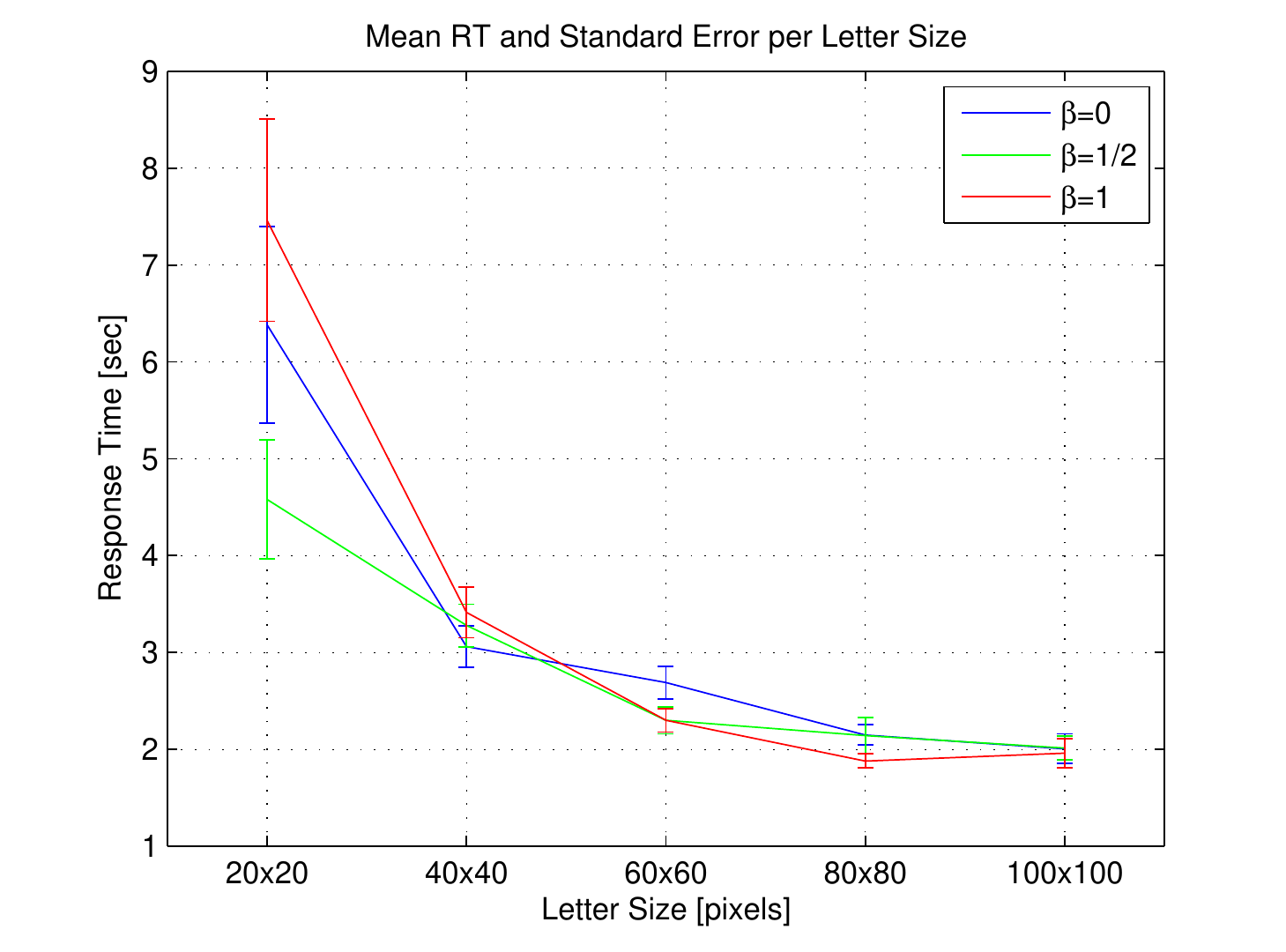} 
\caption[]{Mean Response Time vs. Letter Size in the identification limit test (Experiment 3)}
\label{Fig:Exp3_IT}
\end{figure}

Figure~\ref{Fig:Exp3_IT} presents the mean response time as a function of the letter size.
For all noises, response time decreases with letter size.
Our results therefore coincide with the statement made by \cite{Julesz1964} that time required for stereopsis increases as the size of the hidden object decreases.

In accordance with the results of experiment 2, response time (RT) for pink noise is lower than RT for white noise even when resolution increases. This trend switches direction and indicates superiority of white noise only for letters no larger than $40\times 40$ pixels, i.e. when approaching the identification limits. For such small letters, the amount of correct answers to be accounted for when analyzing the RT is very small, and the observed difference was found to be statistically insignificant.
\newline

We consider the identification limit for letter size in distinguishing between the letters P and B as the minimal size resulting in a correct rate of above 50\%. 
For all the tested noise patterns, the letter size identification limit lies between $20\times 20$ and $40\times 40$ pixels (meaning between 4 and 8 millimeters). 
Assuming the undefinable choice is equivalent to an equally distributed guess, the identification limit for letter size is even lower and can be determined at $20\times 20$ for all the noise patterns.

\section{Conclusions} \label{section:Conclusions}
This work validated the prediction made by \cite{Bruckstein1996} that autostereograms created with pink noise patterns are more easily and correctly perceived than those generated from other noise patterns. 

The first experiment tested how the choice of basic noise pattern affects the time and accuracy of perceiving smooth depth maps. It was found that in accordance with the model of  
\cite{Bruckstein1996}, recognizing smooth depth profiles is easier in autostereograms created from noise patterns with $\beta\geq 1$ than in those with $\beta <1$.
Hence the results indicate that autostereograms created with pink noise patterns exhibit significantly better depth lock-in behavior than those created with white noise and not significantly worse than those created with higher order noises for identifying low resolution objects.

The second experiment checked \cite{Bruckstein1996}'s prediction that recognition of high resolution objects (like letters) will be easy for white and pink noises ($\beta\leq 1$), while harder for smoother noise patterns ($\beta >1$).
It was found that when a mixture of low and high resolution is concerned, response time for pink noise is significantly lower compared with both white and brown noise patterns.
Compensating the contribution of the smooth background to the measured response times based on the first experiment, our results coincide with the prediction of Bruckstein et al.

To further test detection performance for high resolution detail, the third experiment focused on discovering the limit of identifying fine details in the depth dimension.
It should be pointed out that this experiment is preliminary and should be extended. Specifically, an interactive adjustment of the letter size with an adaptive step is expected to give a more accurate estimation of the identification limit for different noise types.
Nevertheless, the experiment clearly proves that even when approaching the high resolution limits, pink noise is comparable to white noise in terms of the identification accuracy.

The three experiments performed convincingly demonstrate the superiority of pink noise based autostereograms over other noise patterns in perception of both low and high resolution objects. 
Hence, our experimental results substantiate the model proposed by \cite{Bruckstein1996} as a good mathematical framework for analyzing autostereograms.

We note in closing that besides the basic noise patterns used for creating the autostereogram, other factors may have an influence on the ease of depth perception in autostereograms.
Future research may, for example, study the effect of using color compared with gray-level SIRDS, using specific colors over others, or using regular structured patterns rather than random noise. While our experiments didn't reveal a clear dependency on the relative depth of the high-resolution detail for superimposed objects, this dependency should also be further studied using a different or wider range of depths.

We believe that autostereograms, beyond being an amazingly popular art form, can continue to help advance the fields of 3D rendering, camouflage, visual physiology and games, and still have considerable stereo-vision research potential.
By validating the theoretical model proposed by \cite{Bruckstein1996}, its underlying assumptions are strengthened, contributing to a better understanding of stereopsis and the correspondence detection mechanism in the human visual system.

\appendix
\section{The Model} \label{sec:appendix}
The model proposed by \cite{Bruckstein1996} for the stereopsis matching process, was developed in order to understand what makes some autostereograms easier to perceive than others.
We here present a simplified one-dimensional problem, referring to each image line individually.

\begin{figure}[htbp] 
\centering \includegraphics[scale=0.4,clip,trim=0cm 0cm 0cm 0cm]{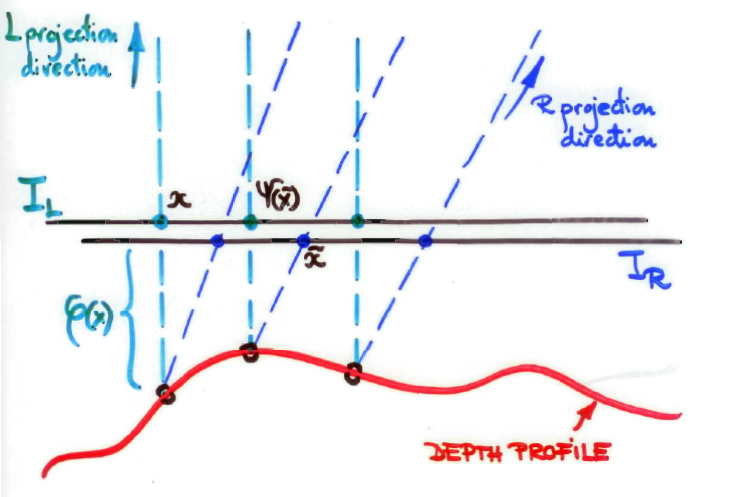}
\caption[]{Stereogram image pairs as seen by the left and right eyes}
\label{Fig:DepthProjection}
\end{figure}

Let us denote the depth profile by $\varphi(x)$.
A general point on the depth surface is projected onto pixel $x$ for the left eye image $I_L$, and on pixel $\tilde{x}$ for the right eye image $I_R$, as illustrated in Figure~\ref{Fig:DepthProjection}. Having the two images fused into one, that means that for the two pixels to be matched as originating from the same point in space, they must have the same value, i.e. $I_L(x)=I_R(\tilde{x})=I(x)$.

The disparity $\Delta$ from $x$ to $\tilde{x}$ is given by
\begin{equation}
\Delta = \frac{E\varphi(x)}{\varphi(x)+D}
\end{equation}
where $E$ is the distance between the eyes and $D$ the viewing distance from the image plane.

Assuming $\varphi(x)$ is bounded, $I$ over any interval $[x,\varphi(x))$ determines $I(\cdot))$ completely.

\begin{figure}[htbp] 
\centering \includegraphics[scale=0.18,clip,trim=0cm 0cm 0cm 0cm]{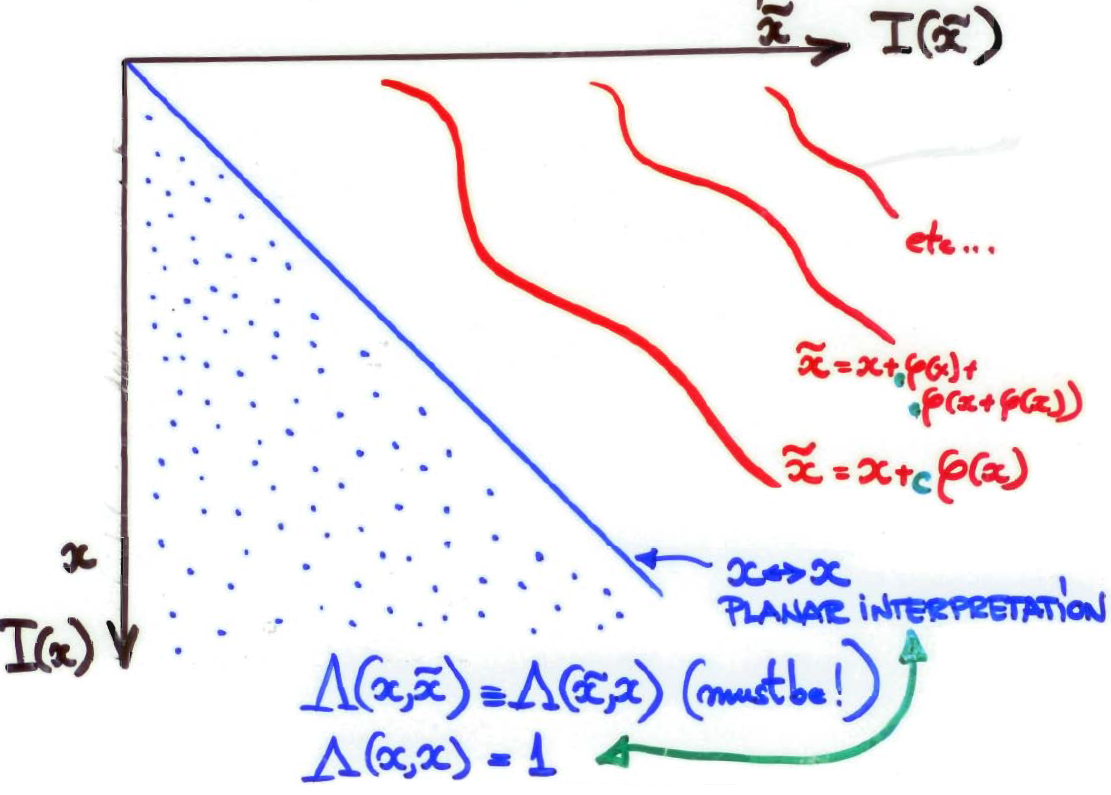}
\caption[]{Illustration of the bivariate matching function}
\label{Fig:BivariateFunction}
\end{figure}

When observing an autostereogram image $I$, the disparities are to be decoded in order to enable perception of the hidden depth dimension. This process could be formulated by defining an abstract bivariate matching function $\Lambda(x,\tilde{x})\in[0,1]$ that indicates how well $I(x)$ locally matches $I(\tilde{x})$ (where a value of 1 indicates a perfect match).
This function has a high ridge along the obvious match (planar interpretation) $x=\tilde{x}$, and additional matches associated with the disparity that occurs in $\tilde{x}=x+\varphi(x)$, $\tilde{x}=x+\varphi(x)+\varphi\left(x+\varphi(x)\right)$, etc.
The ridges of the bivariate matching function are illustrated in Figure~\ref{Fig:BivariateFunction}.

For a better depth lock-in, it is desired to be able to compute a $\Lambda(x,\tilde{x})$ that has a high ridge (i.e. deep basin of attraction) on the desired disparity $\tilde{x}=x+\varphi(x)$, and lower ridges for the other disparities, with valleys between the ridges that enable the interpretation mechanism to move between them reasonably easily.

\begin{figure}[htbp] 
\centering \includegraphics[scale=0.4,clip,trim=0cm 0.5cm 0cm 0.8cm]{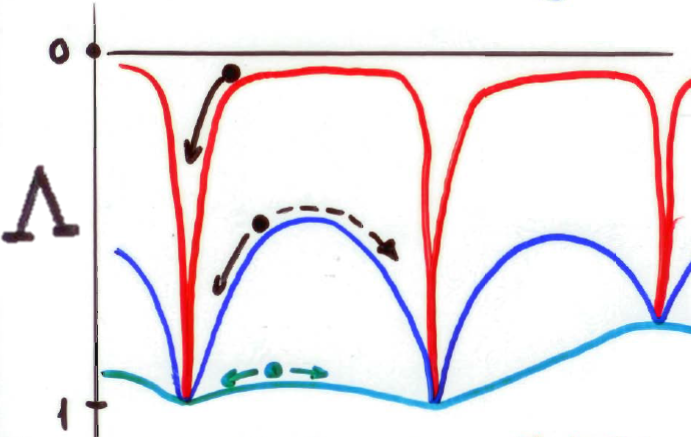}
\caption[]{Matching function's basins of attraction}
\label{Fig:MatchFunction}
\end{figure}

Figure~\ref{Fig:MatchFunction} demonstrates the possible behavior of the matching function for different choices of the basic pattern $I$.
While the red curve has very sharp ridges on the ``correct" disparities, it is difficult to ``leave" the planar interpretation to ``lock in" to the desired depth profile ridge. In the green curve, $I$ leads to blurred ridges and no sharp depth perception.
The blue curve represents the desired function, where the ridges are sharp for the correct disparities, yet the valleys are surmountable.

Embracing the squared difference approach previously suggested by \cite{Sperling1981}, the model of \cite{Bruckstein1996} proposes the matching function to be computed as follows:
\begin{equation}
\Lambda(x,\tilde{x})=f\left(\left[I(x)-I(\tilde{x})\right]^2\right)
\end{equation}
where $f$ is some smooth, monotonically decreasing function satisfying $f(0)=1$ and $\dot{f}(0)<0$.

The sharpness of ridges is here determined by the Laplacian, 
\begin{equation}
\nabla^2\Lambda(x,\tilde{x})\vert_{\tilde{x}=x+\varphi(x)}=2\dot{f}(0)\left\lbrace\left[\frac{d}{dx}I(x)\right]^2+\left[\frac{d}{d\tilde{x}}I(\tilde{x})\right]^2\right\rbrace
\end{equation}
implying that the first derivative of the image controls the shape of the matching function along the disparity line.
In order to create high ridges needed for easy depth perception, one should therefore create an autostereogram with a repeating pattern that has high first-order derivatives in every direction. 
However, we also want few accidental matches (that will obviously occur if $I$ has finite range).
This leads toward the consideration of using random patterns and assuming a matching function based on averaging:
\begin{equation}
\Lambda(x,\tilde{x})=f\left(E\left(\left[I(x)-I(\tilde{x})\right]^2\right)\right)
\end{equation}
where $I(x)$ is defined by extending the rnadom pattern selected on the basic interval $[0,\varphi(0))$.
The averaging process causes false-match peaks to disappear, and so resolves the ambiguity of choosing from multiple possible matches.

The ridge sharpness in the stochastic case is given by
\begin{equation}
\nabla^2\Lambda(x,\tilde{x})=2\dot{f}(0)R''(0)\left\lbrace\left[\frac{d}{dx}B(x)\right]^2+\left[\frac{d}{d\tilde{x}}B(\tilde{x})\right]^2\right\rbrace
\end{equation}
where $B(x)$ is the deterministic ``back-projection" into the interval $[0,\varphi(0))$ and $R(\tau)$ is the auto-correlation of the process $I(x)$.

This result suggests we can control the shape of the ridges by choosing a random process as the basic pattern with high second derivative at the origin. A completely uncorrelated random process such as white noise seems ideal in that respect. However, the basin of attraction with a completely uncorrelated random process for the obvious match could be so deep that it will be hard to direct the visual system to the second ridge of depth encoding disparities.

Here \cite{Bruckstein1996} adopts a coarse-to-fine model as in \cite{Marr1979}, assuming that images presented to us are filtered by several low-pass, or band-pass, filters to create a pyramid of coarser and coarser images, and the perceptual system works its way from coarse to fine scale to perceive depth in various resolutions. So to perceive depth optimally, the random process we use as the basic pattern of the autostereograms needs be scale invariant. Furthermore, because we aim to direct the visual system from coarser to finer resolution, it is a desired attribute that the basins of attraction at each level will get narrower at the finer resolutions.

Let us look at the autocorrelation $R(\tau)$ of some random process with power spectral density $S(f)$:
\begin{equation}
R(\tau)=\int_0^\infty S(f)e^{j2\pi f\tau}df
\end{equation}
\begin{equation}
\frac{d^2R(\tau)}{d\tau^2}=-4\pi\int_0^\infty f^2S(f)e^{j2\pi f\tau}df
\end{equation}

If we use a low pass filter with a cut-off frequency of $f_0=\frac{1}{\sigma}$, and normalize $\tau$ by $\sigma$, we get
\begin{equation}
\frac{d^2R_\sigma(\tau)}{d(\frac{\tau}{\sigma})^2}=
\sigma^2\frac{d^2R_\sigma(\tau)}{d\tau^2}=
-4\pi\sigma^2\int_0^{\frac{1}{\sigma}} f^2S(f)e^{j2\pi f\tau}df
\end{equation}

For noise patterns with power spectra of the form $S(f)=Cf^{-\beta}$ ($\beta\in\mathbb{R}$, $C$ constant):
\begin{equation}
R_\sigma''(0)=-4\pi C\sigma^2\int_0^{\frac{1}{\sigma}} f^{2-\beta}df=\frac{-4\pi C}{3-\beta}\sigma^{\beta-1}
\end{equation}
for $\beta\neq 3$.

It is readily observed that for $\beta=1$ (pink noise), $R_\sigma''(0)$ is independent of $\sigma$, meaning the peaks of the matching function have constant normalized width in scale-space, giving the desired scale invariance properties.
From an unnormalized point of view, the peaks get narrower with scale from coarse to fine, as intended. 

Therefore the model predicts that pink noise leads to easy depth lock-in across scales with excellent detail perception
\newline

To demonstrate this concept using our generated autostereograms, we averaged the result of $[I(x)-I(\tilde{x})]^2$ over a chosen subset of successive image lines in several images generated with the same depth profile and the same noise type.
The matching function is then calculated as $\Lambda(x,\tilde{x})=f\left(E\left(\left[I(x)-I(\tilde{x})\right]^2\right)\right)$ for $f(z)=\frac{1}{1+\lambda z}$, with $\lambda=0.001$ (chosen empirically).

\begin{figure*}[htbp]
\centering 
\subfloat[]{
\centering \includegraphics[scale=0.22,clip,trim=0cm 0cm 0cm 1cm]{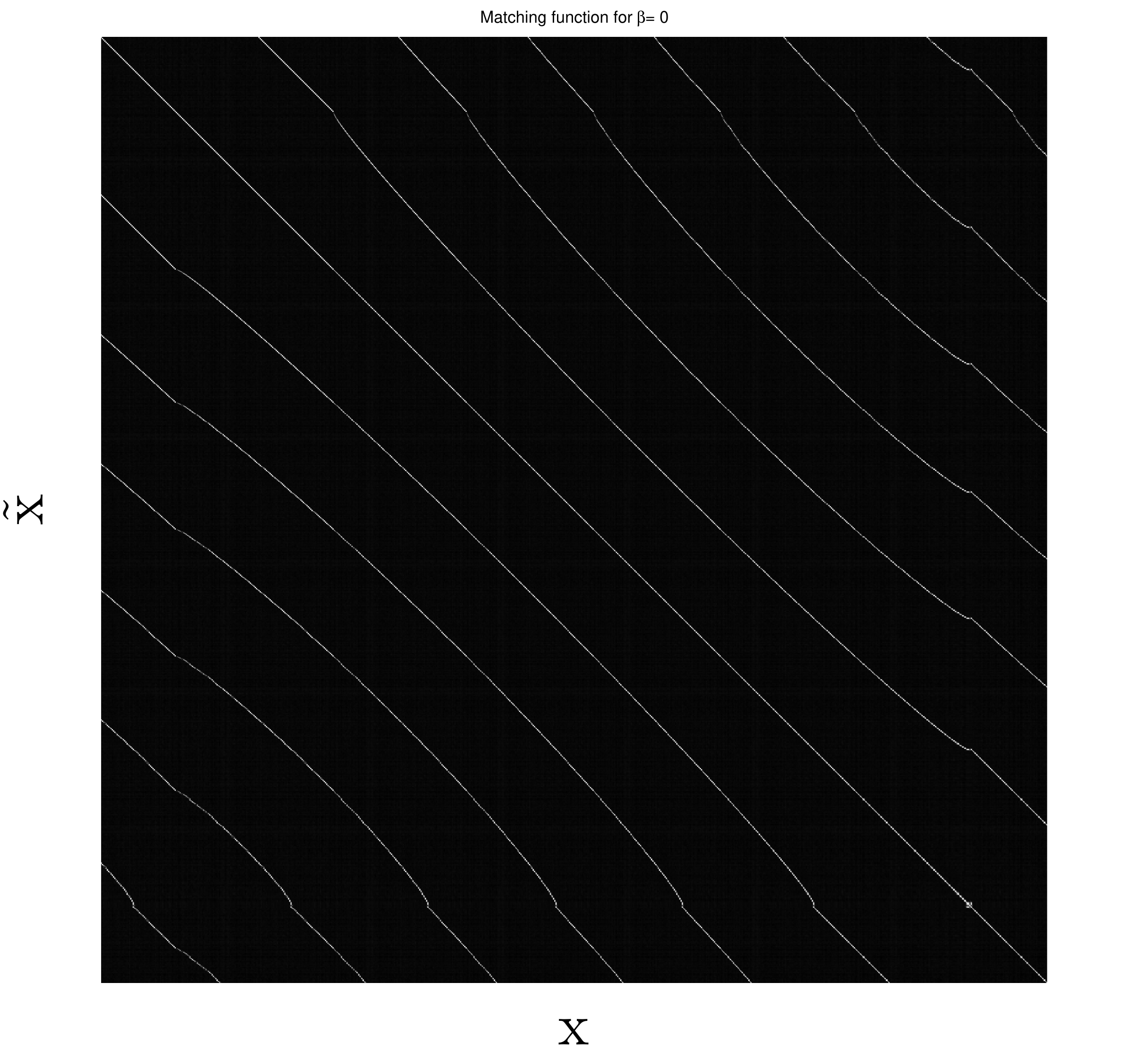}
\label{Fig:match0a}
}
\subfloat[]{
\centering \includegraphics[scale=0.45,clip,trim=0cm 0cm 0cm 0.5cm]{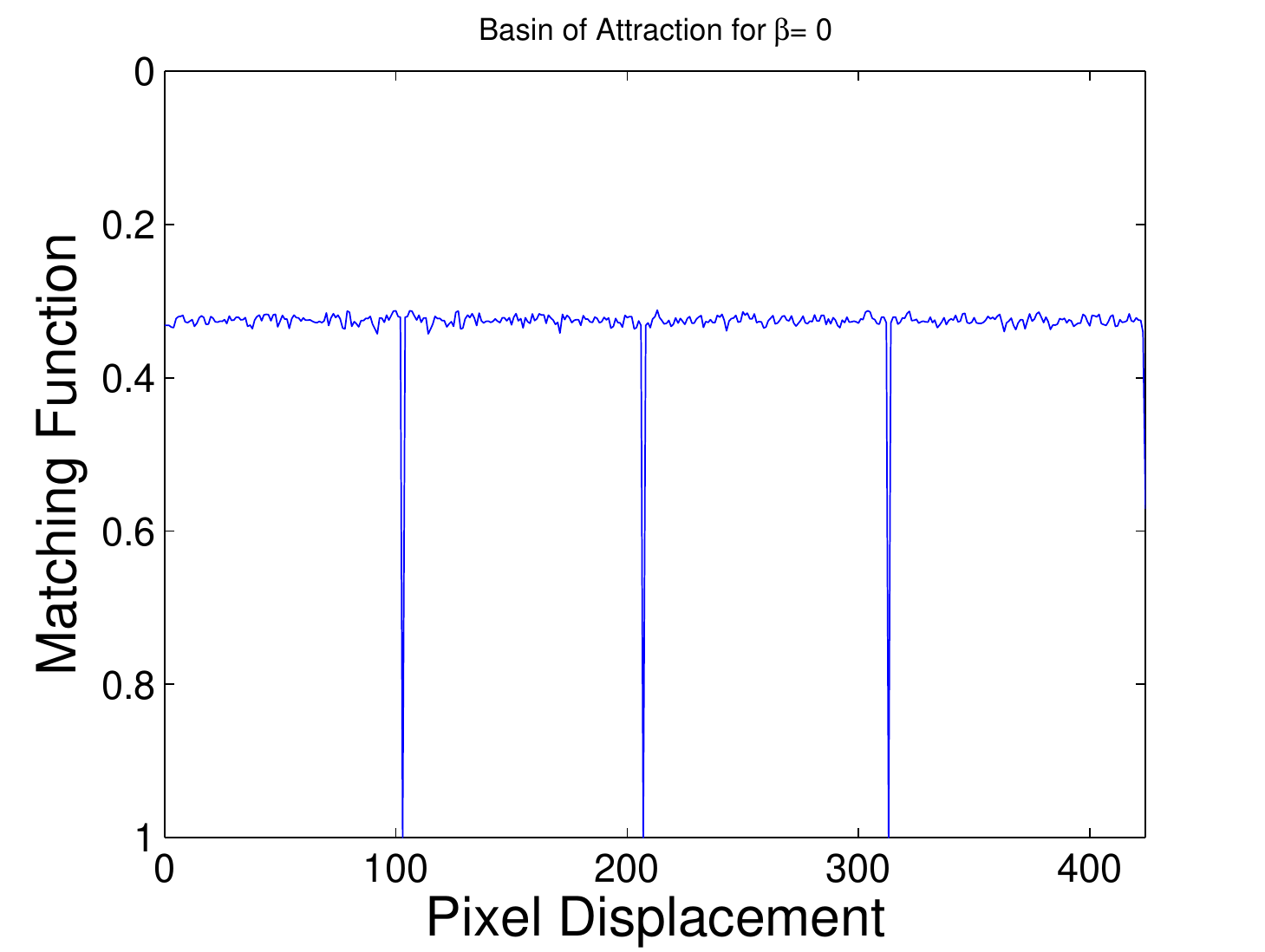}
\label{Fig:match0b}
}
\\
\subfloat[]{
\centering \includegraphics[scale=0.22,clip,trim=0cm 0cm 0cm 1cm]{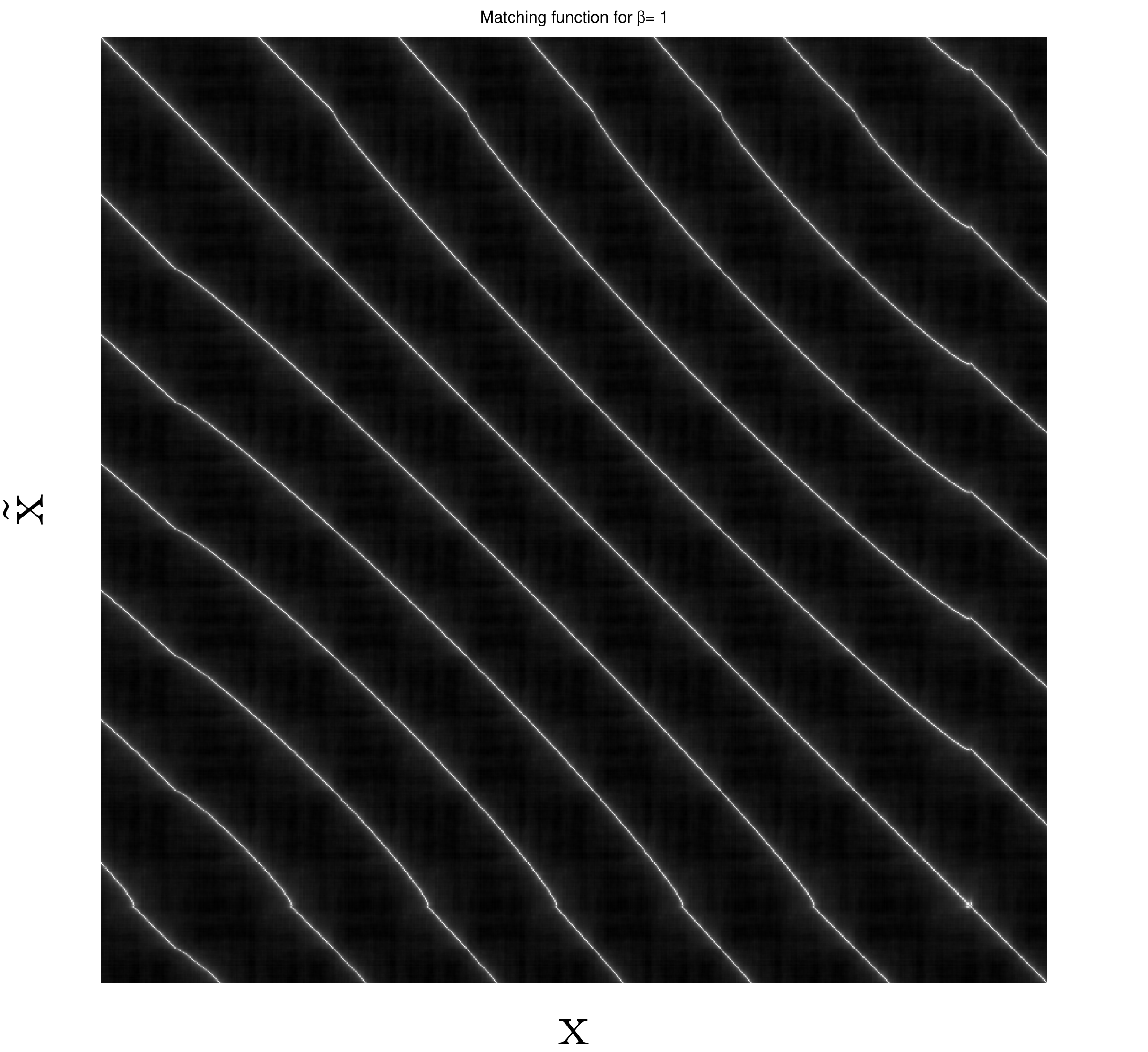}
\label{Fig:match1a}
}
\subfloat[]{
\centering \includegraphics[scale=0.45,clip,trim=0cm 0cm 0cm 0.5cm]{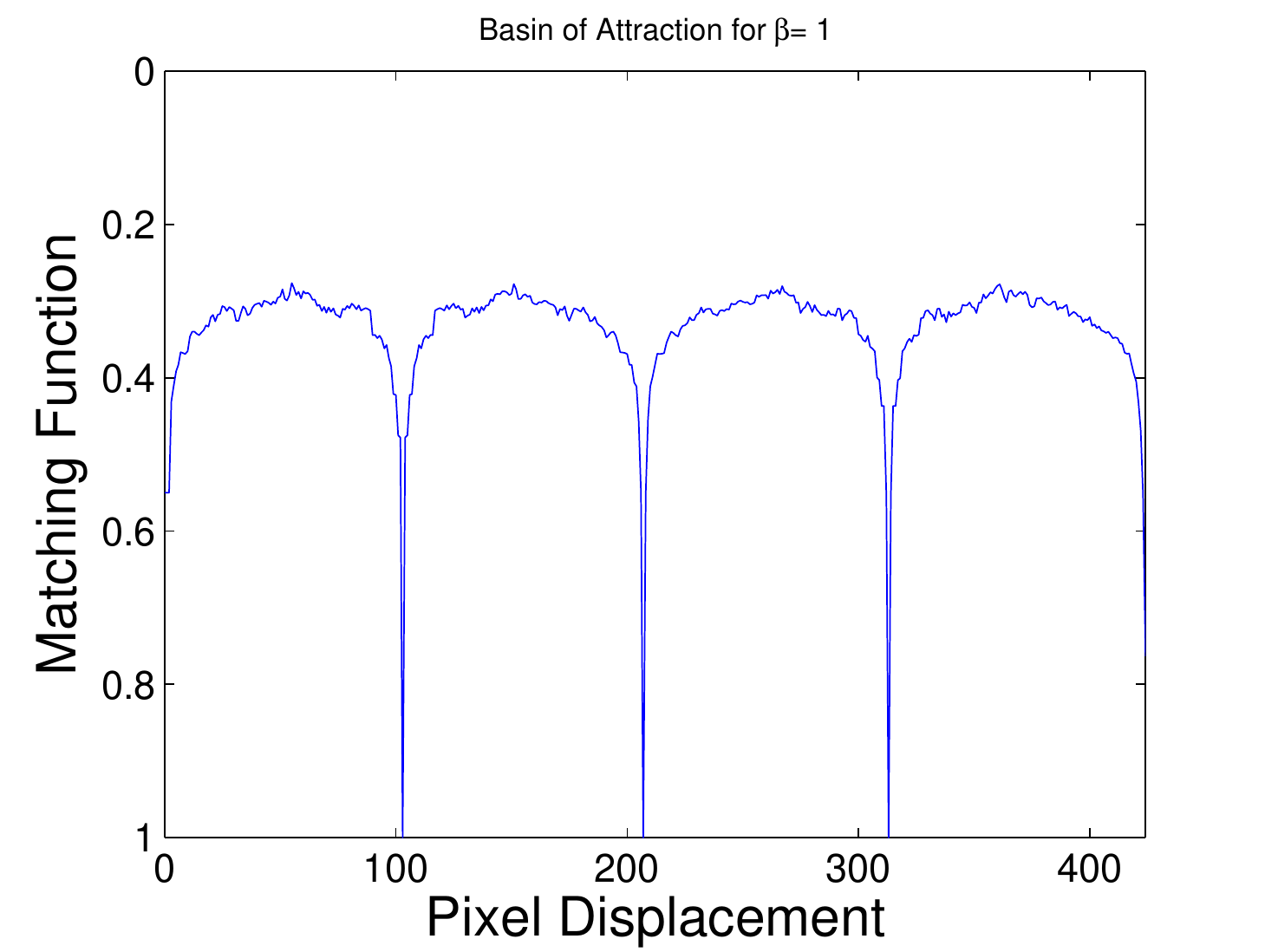}
\label{Fig:match1b}
}
\\
\subfloat[]{
\centering \includegraphics[scale=0.22,clip,trim=0cm 0cm 0cm 1cm]{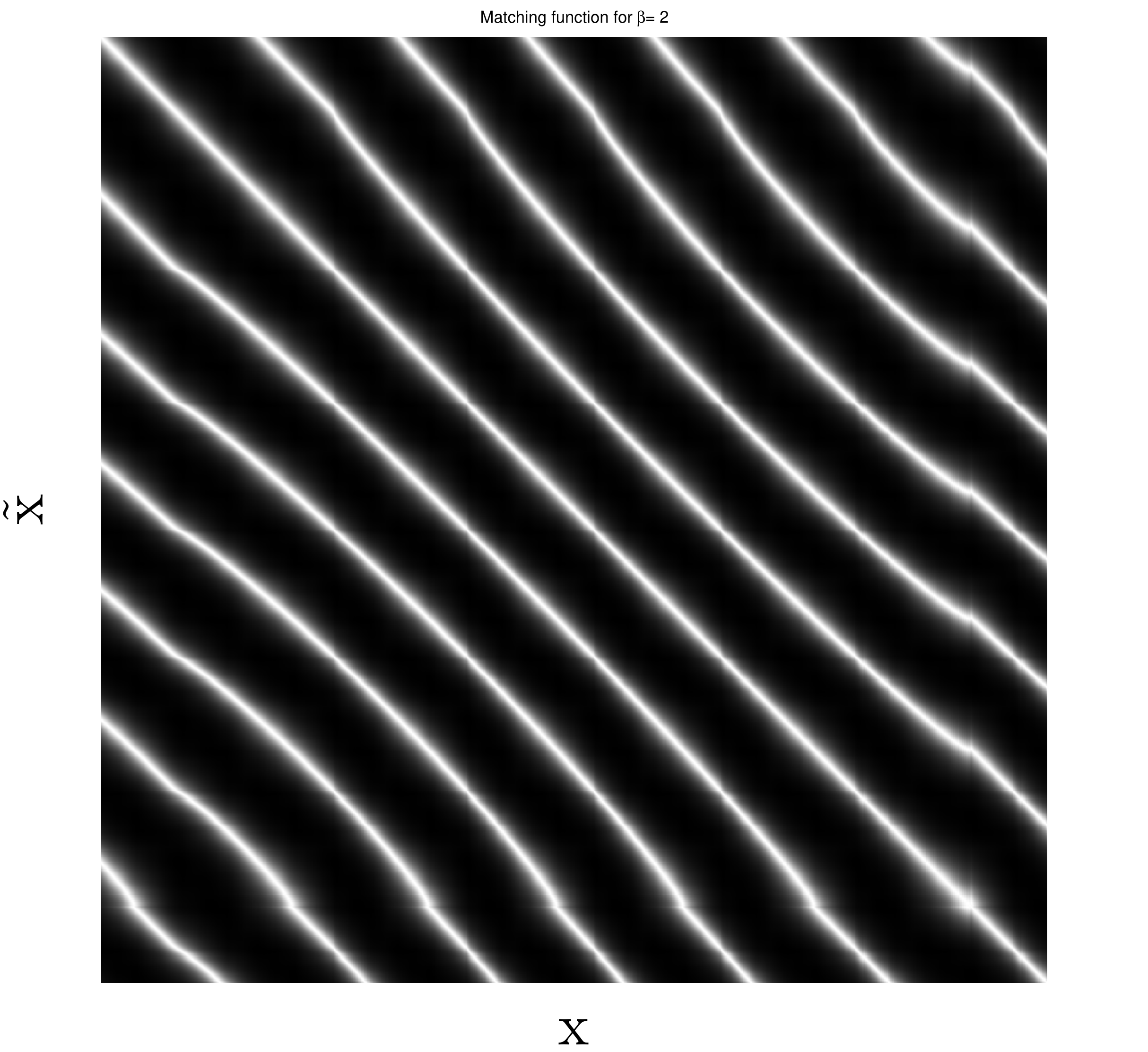}
\label{Fig:match2a}
}
\subfloat[]{
\centering \includegraphics[scale=0.45,clip,trim=0cm 0cm 0cm 0.5cm]{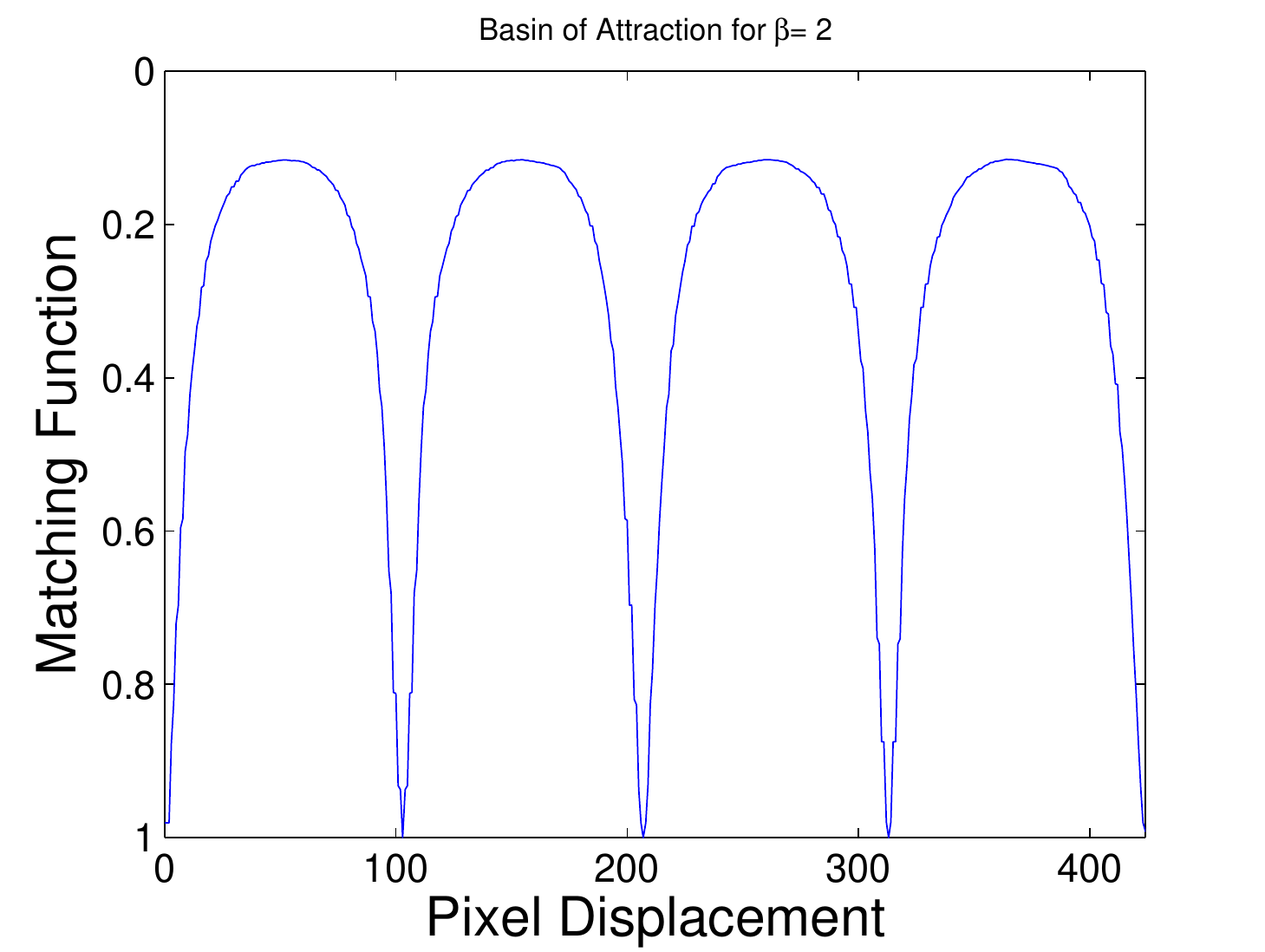}
\label{Fig:match2b}
}

\caption[]{Matching function $\Lambda(x,\tilde{x})$ (left) and basins of attraction (right) for \subref{Fig:match0a}-\subref{Fig:match0b} White noise, \subref{Fig:match1a}-\subref{Fig:match1b} Pink noise, \subref{Fig:match2a}-\subref{Fig:match2b} Brown noise. The basins of attraction are displayed as a function of the pixel displacement with respect to the planar interpretation $\tilde{x}=x$. 
}
\label{Fig:MatchFunctionNoises}
\end{figure*}

The result of those calculations for autostereograms created with white, pink and brown noise patterns are presented in Figure~\ref{Fig:MatchFunctionNoises}.
We also present the counter-diagonal of the matching function to show a one-dimensional view of the basins of attraction.

It can be observed that for white noise, the matching function has very sharp ridges, but moving between ridges is quite difficult due to the difficulty in getting out from the planar interpretation ridge.
For brown noise, the matching function has wider ridges and smooth valleys, which allows for an easier movement of interpretation between ridges, but may mean a blurry image.
Pink noise represents the balance between them, having sharp ridges and valleys that are surmountable.

\pagebreak

\bibliographystyle{plain} 
\bibliography{as_refs}

\begin{thebibliography}{10}

\bibitem{Arndt1995}
Petra~A. Arndt, Hanspeter~A. Mallot, and Heinrich~H. B\"ulthoff.
\newblock {Human Stereovision without Localized Image Features}.
\newblock {\em Biol. Cybern}, 72:279--293, 1995.

\bibitem{Banks2004}
M.~S. Banks, S.~Gepshtein, and M.~S. Landy.
\newblock {Why is Spatial Stereoresolution So Low?}
\newblock {\em Journal of Neuroscience}, 24(9):2077--2089, March 2004.

\bibitem{Brainard1997}
D.~H. Brainard.
\newblock The {P}sychophysics {T}oolbox.
\newblock {\em Spatial Vision}, 10(4):433--436, 1997.

\bibitem{Brewster1844}
David Brewster.
\newblock {On the Knowledge of Distance given by Binocular Vision}.
\newblock {\em Transactions of the Royal Society of Edinburgh}, 15:663--675,
  1844.

\bibitem{Bruckstein1996}
A.~M. Bruckstein, R.~Onn, and T.~J. Richardson.
\newblock {Improving the Vision of Magic Eyes: A Guide to Better
  Autostereograms}.
\newblock In {\em Advances in Image Understanding}, pages 158--176. IEEE
  Computer Society Press, Los Alamitos, 1996.

\bibitem{Burge2014}
J.~Burge and W.~S. Geisler.
\newblock {Optimal Disparity Estimation in Natural Stereo Images}.
\newblock {\em Journal of Vision}, 14(2), 2014.

\bibitem{Burton1987}
G.~J. Burton and Ian~R. Moorhead.
\newblock {Color and Spatial Structure in Natural Scenes}.
\newblock {\em Applied Optics}, 26(1):157--170, 1987.

\bibitem{Ditzinger2000}
T.~Ditzinger, M.~Stadler, D.~Str\"uber, and J.~A.~S. Kelso.
\newblock {Noise Improves Three-Dimensional Perception: Stochastic Resonance
  and Other Impacts of Noise to the Perception of Autostereograms}.
\newblock {\em Phys. Rev. E}, 62(2):2566--2575, 2000.

\bibitem{Geselowitz2003}
L.~Geselowitz.
\newblock {The AbSIRD Project: To Create Real-Time SIRDs}, 2003.

\bibitem{Gomez2012}
Aurora~Torrents G\'{o}mez, N\'{u}ria Lup\'{o}n, Gen\'{i}s Cardona, and
  J~Antonio Aznar-Casanova.
\newblock {Visual Mechanisms Governing the Perception of Auto-stereograms}.
\newblock {\em Clin Exp Optom}, 95:146--152, 2012.

\bibitem{Harris1997}
J.~M. Harris, S.~P. McKee, and H.~S. Smallman.
\newblock {Fine-Scale Processing in Human Binocular Stereopsis}, volume = {14},
  month = {Aug}, year = {1997},.
\newblock {\em J. Opt. Soc. Am. A}, (8):1673--1683.

\bibitem{Ittelson1960}
W.~H. Ittelson.
\newblock {\em {Visual Space Perception}}.
\newblock Springer Publishing Co., New York, 1960.

\bibitem{Julesz1964}
B.~Julesz.
\newblock {Binocular Depth Perception without Familiarity Cues}.
\newblock {\em Science}, 145(3630):356--362, 1964.

\bibitem{Kimmel2002}
Ron Kimmel.
\newblock {3D Shape Reconstruction from Autostereograms and Stereo}.
\newblock {\em J. Visual Communication and Image Representation},
  13(1-2):324--333, 2002.

\bibitem{Kleiner2007}
M.~Kleiner, D.~Brainard, and D.~Pelli.
\newblock {What's New in Psychtoolbox-3?}
\newblock In {\em Perception}, volume~36. ECVP Abstract Supplement, 2007.

\bibitem{Lau2002}
Mark S.~K. Lau and C.~P. Kwong.
\newblock {Analysis of Echoes in Single-Image Random-Dot-Stereograms}.
\newblock {\em J. Math. Imaging Vis.}, 16(1):69--79, 2002.

\bibitem{Li1994}
Z.~Li and J.~J. Atick.
\newblock {Efficient Stereo Coding in the Multiscale Representation}.
\newblock {\em Network: Computation in Neural Systems}, 5(2):157--174, 1994.

\bibitem{Marr1979}
D.~Marr and T.~Poggio.
\newblock {A Computational Theory of Human Stereo Vision}.
\newblock {\em Proceedings of the Royal Society of London. Series B. Biological
  Sciences}, 204(1156):301---328, 1979.

\bibitem{Minh2002}
S.~T. Minh, K.~Fazekas, and A~Gschwindt.
\newblock {The Presentation of Three-Dimensional Objects with Single Image
  Stereogram}.
\newblock {\em IEEE Transactions on Instrumentation and Measurement},
  51(5):955--961, 2002.

\bibitem{Minh2001}
S.~T. Minh, G.~Marosi, and A.~Gschwindt.
\newblock {Case Study of Autostereoscopic Image Based on SIRDS Algorithm }.
\newblock {\em Periodica Polytechnica Ser. El. Eng.}, 45(2):119--138, 2001.

\bibitem{Magic1993}
{N. E. Thing Enterprises}.
\newblock {\em {Magic Eye: A New Way of Looking at the World}}.
\newblock Michael Joseph LTD, Penguin Group, 1993.

\bibitem{Palmer1999}
S.~E. Palmer.
\newblock {\em {Vision Science: Photons to Phenomenology}}.
\newblock A Bradford book. Massachusetts Institute of Technology, 1999.

\bibitem{Qian1997}
N.~Qian and Y.~Zhu.
\newblock Physiological computation of binocular disparity.
\newblock {\em Vision Research}, 37(13):1811--1827, 1997.

\bibitem{Read2007}
J.~C. Read and B.~G. Cumming.
\newblock {Sensors for Impossible Stimuli May Solve the Stereo Correspondence
  Problem}.
\newblock {\em {Nature Neuroscience}}, 10(10):1322--1328, October 2007.

\bibitem{Reimann1995}
D.~Reimann, T.~Ditzinger, E.~Fischer, and H.~Haken.
\newblock {Vergence Eye Movement Control and Multivalent Perception of
  Autostereograms}.
\newblock {\em Biological Cybernetics}, 73(2):123--128, 1995.

\bibitem{Sperling1981}
G.~Sperling.
\newblock {Mathematical Models of Binocular Vision}.
\newblock In S.~Grossberg, editor, {\em {SIAM-AMS proceedings}}, volume~13,
  pages 281--300, 1981.

\bibitem{Terrell1994}
Maria~S. Terrell and Robert~E. Terrell.
\newblock {Behind the Scenes of a Random Dot Stereogram}.
\newblock {\em The American Mathematical Monthly}, 101(8):715--724, 1994.

\bibitem{Thimbleby1994}
Harold~W. Thimbleby, Stuart Inglis, and Ian~H. Witten.
\newblock {Displaying 3D Images: Algorithms for Single-Image Random-Dot
  Stereograms}.
\newblock {\em Computer}, 27(10):38--48, 1994.

\bibitem{Torralba2003}
A.~Torralba and A.~Oliva.
\newblock {Statistics of Natural Image Categories}.
\newblock {\em Network: Computation in Neural Systems}, 14:391--412, 2003.

\bibitem{Tyler1977}
C.~W. Tyler and J.~J. Chang.
\newblock Visual echoes: The perception of repetition in quasi-random patterns.
\newblock {\em Vision Research}, 17(1):109--116, 1977.

\bibitem{Tyler1990}
Christopher~W. Tyler and Maureen~B. Clarke.
\newblock {The Autostereogram}.
\newblock In {\em {Proc. SPIE}}, volume 1256, pages 182--197, 1990.

\bibitem{Wilmer2008}
Jeremy~B Wilmer and Benjamin~T Backus.
\newblock {Self-Reported Magic Eye Stereogram Skill Predicts Stereoacuity}.
\newblock {\em Perception}, 37(8):1297--1300, 2008.

\end{thebibliography}

\end{document}